\documentclass{article}

\usepackage{microtype}
\usepackage{graphicx}
\usepackage{subfigure}
\usepackage{booktabs} 

\usepackage{hyperref}

\usepackage[accepted]{icml2023}

\usepackage{amsmath}
\usepackage{amssymb}
\usepackage{mathtools}
\usepackage{amsthm}

\usepackage[capitalize,noabbrev]{cleveref}

\theoremstyle{plain}
\newtheorem{theorem}{Theorem}[section]

\theoremstyle{definition}
\newtheorem{definition}[theorem]{Definition}

\theoremstyle{remark}
\newtheorem{remark}[theorem]{Remark}

\usepackage[textsize=tiny]{todonotes}

\usepackage{listings}

\icmltitlerunning{Continuous Spatiotemporal Transformers}

\begin{document}

\twocolumn[
\icmltitle{Continuous Spatiotemporal Transformer}



\icmlsetsymbol{equal}{*}

\begin{icmlauthorlist}
\icmlauthor{Antonio Henrique de Oliveira Fonseca}{dpt1}
\icmlauthor{Emanuele Zappala}{dpt2}
\icmlauthor{Josue Ortega Caro}{dpt3,dpt6}
\icmlauthor{David van Dijk}{dpt2,dpt4,dpt5,dpt6}
\end{icmlauthorlist}

\icmlaffiliation{dpt1}{Interdepartmental Neuroscience Program,
    Yale University, New Haven, CT, USA}
\icmlaffiliation{dpt2}{Department of Computer Science,
	Yale University, New Haven, CT, USA}
 \icmlaffiliation{dpt3}{Department of Neuroscience,
    Yale University, New Haven, CT, USA}
 \icmlaffiliation{dpt4}{Department of Internal Medicine (Cardiology),
    Yale University, New Haven, CT, USA}
 \icmlaffiliation{dpt5}{Interdepartmental Program in Computational Biology \& Bioinformatics, Yale University, New Haven, CT, USA}
  \icmlaffiliation{dpt6}{Wu Tsai Institute, Yale University, New Haven, CT, USA}

\icmlcorrespondingauthor{David van Dijk}{david.vandijk@yale.edu}


\icmlkeywords{Machine Learning, ICML}

\vskip 0.3in
]



\printAffiliationsAndNotice{}  

\begin{abstract}
Modeling spatiotemporal dynamical systems is a fundamental challenge in machine learning. Transformer models have been very successful in NLP and computer vision where they provide interpretable representations of data. However, a limitation of transformers in modeling continuous dynamical systems is that they are fundamentally discrete time and space models and thus have no guarantees regarding continuous sampling. To address this challenge, we present the Continuous Spatiotemporal Transformer (CST), a new transformer architecture that is designed for modeling of continuous systems. This new framework guarantees a continuous and smooth output via optimization in Sobolev space. We benchmark CST against traditional transformers as well as other spatiotemporal dynamics modeling methods and achieve superior performance in a number of tasks on synthetic and real systems, including learning brain dynamics from calcium imaging data. 

\end{abstract}

\section{Introduction}
\label{introduction}

\begin{figure}[t]
\centering
  \includegraphics[width=0.9\columnwidth]{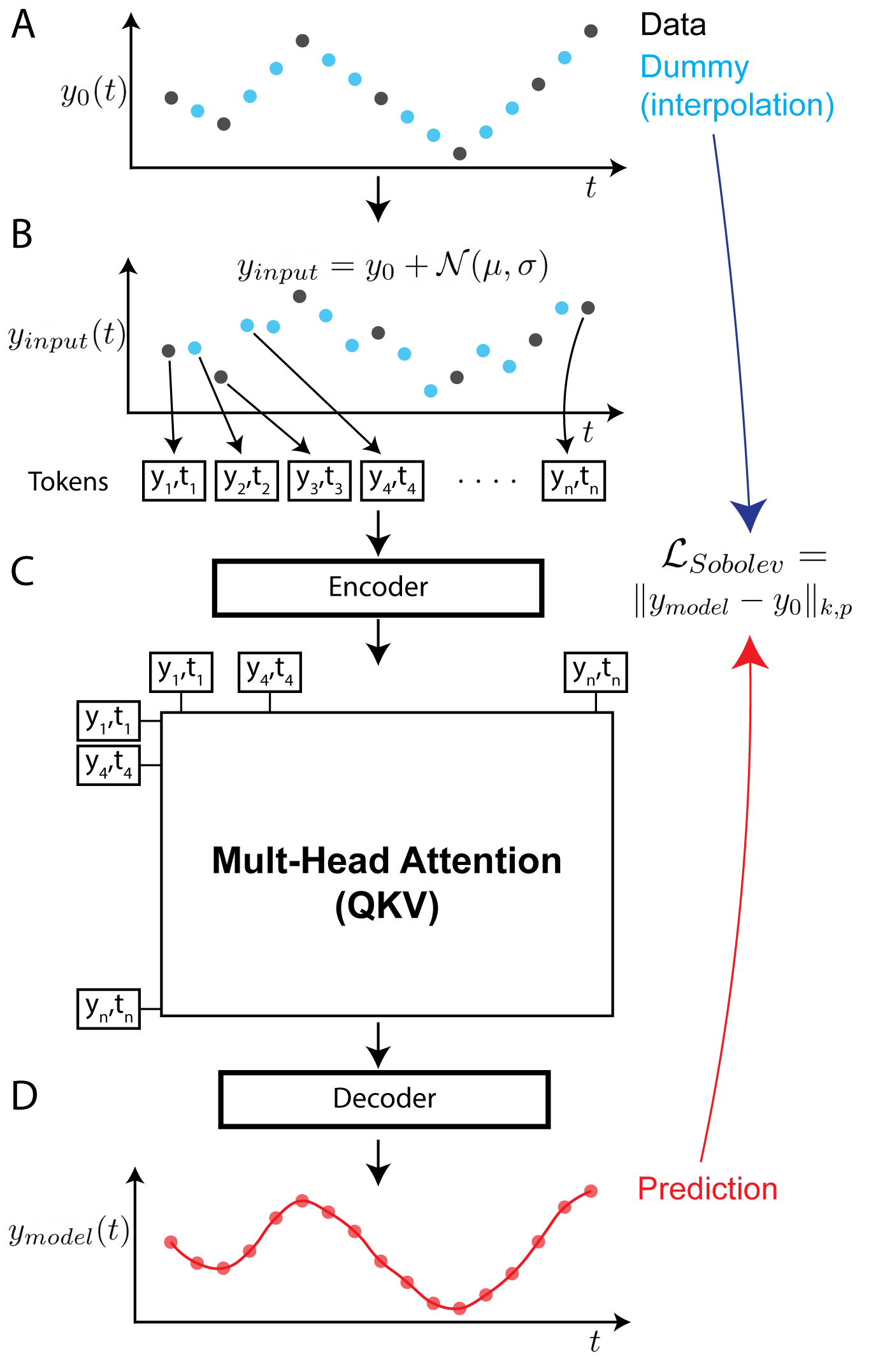}
  \caption{Diagram of CST's workflow. (A) The model receives a mix of real and "dummy" data points. These points are initialized via a linear interpolation of the real data points. (B) All points are perturbed with Gaussian noise. (C) Each point is treated as a token of the sequence. The points and their positional information are encoded to a latent space and fed to a multi-head self-attention module. (D) The model's output is a prediction for each input coordinate. The model is trained to minimize the Sobolev loss.}
  \label{fig:method}
\end{figure}

The theory of dynamical systems has found profound applications throughout the sciences, both theoretical and applied. Traditionally, dynamical system analysis aims to find the rules that govern the dynamics of an underlying system. In this setting, we first obtain a model that describes the given system, either through theoretical principles (model-based) or through experimental data (data-driven) \citep{ghadami2022data}, and then study the mathematical properties of it.
Having a model of the dynamical system grants a deeper understanding of the phenomena, allowing for predictions of the system state continuously in time \citep{ghadami2022data,krishnapriyan2022learning}. Such dynamical systems can be found throughout engineering and science. In biology, the brain is a notably complex dynamical system \citep{wang2016brain}. The spiking activity of neurons within the neural population produces complex spatiotemporal patterns \citep{muller2018cortical}.  These neural activity patterns represent a dynamical system that evolves over time, where the state of the system is defined in terms of the joint firing patterns of the neural populations \citep{vyas2020computation}. For complex systems such as the brain, model-based approaches for learning dynamics are not amenable, thus the dynamics have to be learned directly from collected data. However, learning continuous dynamics from discretely sampled data is challenging and is an active area of study in machine learning \citep{willard2020integrating}. 

Among the continuous time model approaches, Neural ODEs \cite{chen2018neural, rubanova2019latent} have found important applications. While effective in modeling temporal dynamics, these models are unable to capture long-range spatiotemporal relations in the data and do not provide interpretable models  \cite{zappala2022neural}. In the meantime, Transformers \citep{vaswani2017attention} have become state-of-the-art in several tasks across domains \citep{lu2021pretrained}, wherein their performance is mainly attributed to their capacity to capture long-range dependencies in the data as well as their training scalability \citep{bertasius2021space}.

Despite their widespread use across domains \citep{lu2021pretrained}, Transformers are still restricted to discrete space and time applications. In this work, we demonstrate that the standard Transformer has limitations in modeling continuous systems. To address this, we introduce new regularizations to the Transformer architecture, resulting in a new framework hereafter called \textbf{Continuous Spatiotemporal Transformer (CST)}. We show that CST is capable of modeling continuous data, resulting in smooth output and good interpolation performance. To showcase our method, we first validate CST on a toy-dataset and different popular benchmark datasets. Finally, we test CST in modeling dynamics in neural experimental data. We compare CST to other commonly used methods for modeling sequential data in three tasks: 1) modeling 2D spirals generated by integral equations (Sec. \ref{sec:2d_spirals}); 2) modeling a benchmark video-dataset (KITTI) (Sec. \ref{sec:benchmark_video}); 3) modeling fluid dynamics (Sec. \ref{sec:NS_experiment}); and finally 4) extracting behaviorally meaningful latent representations of the dynamics from widefield calcium imaging recordings. 


We summarize our contributions as follows:
\begin{itemize}
\item We show the limitations of Transformers in modeling continuous data.
\item We introduce a new framework that allows the application of Transformers to continuous systems.
\item We show that our method provides accurate interpolation of both data and attention weights.
\item Finally, we use our method to model brain activity recordings and show that the attention weights encode meaningful information about the dynamics. 
\end{itemize}

\begin{figure*}[t]
\centering
  \includegraphics[width=\textwidth]{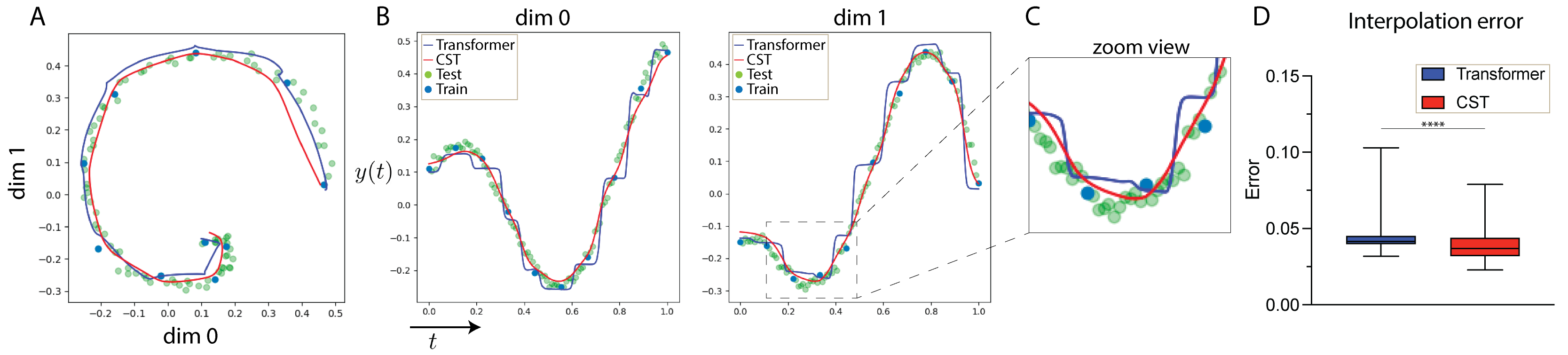}
  \caption{Continuous sampling of Transformer and CST. The Transformer shows step-like behavior whereas CST is smooth. A) Example of model fits to 2D spirals. B) Individual spiral dimensions over time. Both CST and the Transformer were trained to fit the data sampled from the spirals ('Train' blue points). During inference, the models were evaluated at 1000 coordinates along the spiral (lines shown for CST (red) and the Transformer (blue)) (see Figure \ref{fig:2d_spiral_moreExamples} for more examples). C) Zoomed-in view, emphasizing the difference in smoothness between CST and the Transformer.  D) The interpolation error (L2-norm) for the test points (green) shows that CST has significantly ($P<0.0001$) better interpolation than the Transformer.}
  \label{fig:2D_IE_curves_CST_vs_Transformers}
\end{figure*}
 

\section{Background and Related work}\label{sec:background}

\subsection{Operator learning}

 Learning operators, i.e. mappings between function spaces, is a machine learning task with fundamental applications to spatiotemporal dynamical systems \cite{kovachki2021neural, zappala2022neural,cao2021choose}. In fact, several dynamical systems are modeled through ordinary differential equations (ODEs), partial differential equations (PDEs), or integral equations (IEs), and one is interested in finding the operators corresponding to the equations governing the dynamics. However, in practice, we do not have a mathematical model that describes the behavior of the system, but we rather have data sampled from instances of the system. In such circumstances, we are interested in learning an operator that corresponds to the system. An example could be learning an operator that maps the function representing a system at time $t=0$ to the system at later time points.
 This is the setting of operator learning problems, and several approaches, including using deep learning, have been presented \citep{kovachki2021neural,lu2021learning,li2020fourier,li2020neural,cao2021choose}. Operator learning problems are often formulated on finite grids, and passing to the continuous limit is a significant issue. Moreover, in practical cases such as in physics, it is of interest to be able to compute derivatives of the model's output, in which case smoothness is needed. 
 Our main goal and contribution in this article are to introduce an operator learning framework for continuous and smooth functions on space-time domains.

 \subsection{Sobolev spaces}
 
 Sobolev spaces were introduced as a framework for solving differential equations \citep{Brezis}. In such spaces, one studies weak solutions for differential equations, i.e. solutions that hold almost everywhere with respect to the integral over the domain of interest. Then, for regular enough solutions, the equation is also a strong solution, i.e. a solution in the usual sense where equality holds for each point of the domain without the sign of integral. The study of such spaces also leads to the notion of weak differentiability and the Sobolev norm, which is a norm that takes into account the function itself as well as its derivatives. Sobolev spaces are a fundamental object of study in functional analysis, especially in relation to differential equation theory. More recently, they have found important applications in machine learning, where minimizing the Sobolev norm with respect to target data as well as its derivatives has shown good regularization effects \citep{czarnecki2017sobolev,son2021sobolev,kissel2020sobolev,cardona2022replacing,fischer2020sobolev,vlassis2021sobolev}. Our optimization task is formulated in the Sobolev space to ensure that the learned operator outputs functions that are both continuous and smooth. However, our approach differs with respect to previous methods in that we do not use the derivatives of the target data functions explicitly, but we rather minimize the $p$-norm of the higher derivatives on sampled points, without directly comparing them to data. Therefore, our approach does not require extra knowledge or computation of the derivatives of data.
 
\subsection{Continuous time models}

A fundamental issue in machine learning is that of modeling continuous systems from discretely sampled data. Mathematical modeling of dynamical systems in the sciences and engineering, in fact, is performed through continuous and differentiable functions, due to their favorable analytical properties. We are therefore interested in machine learning models whose output is continuous and smooth, and that can therefore be interpolated with accuracy even when the data set is irregularly sampled. Several methods have been proposed, e.g. \cite{chen2018neural,rubanova2019latent,poli2020hypersolvers,zappala2022neural}, based on the idea of solvers. In contrast, our approach combines operator learning techniques based on transformers \cite{vaswani2017attention} and Sobolev norm \cite{Brezis} to obtain an operator that outputs smooth functions with a high degree of accuracy on interpolation tasks for irregularly sampled data. 

\subsection{Transformers}
The self-attention mechanism and Transformers models were introduced in \citet{vaswani2017attention} and have shown exquisite performance in sequence modeling problems, such as natural language processing (NLP). Since its first appearance, Transformers have excelled in several domains \citep{lu2021pretrained}. The Transformer uses self-attention mechanisms to learn the relationship between the elements of a sequence and use this information to make contextualized predictions. When trained on large corpora, Transformers can learn to abstract semantics from the text \citep{devlin2018bert}. The state-of-the-art performance of Transformers in NLP is attributed to their capacity to capture long-range dependencies among words (i.e. extract contextual meaning) as well as their training scalability \citep{bertasius2021space}. More recently, studies focused on the computation performed by self-attention have shown it acts as a learnable integral kernel with non-local properties. This makes the Transformer especially fit for learning complex sequential data with long-range dependencies \citep{cao2021choose,cao2022understand}, while also being computationally efficient for long sequences \citep{choromanski2020rethinking}.


\subsection{Modeling brain dynamics}

Modeling brain dynamics has been a focal point of neuroscience since its start \cite{hodgkin1952quantitative,rall1959branching}. However, until recently, technological limitations have significantly hindered the field in two perspectives: 1) Difficulties in collecting high-throughput data, and 2) computational limitations to model complex non-linear dynamics \cite{stevenson2011advances}. Recently, several neural-network-based methods have been developed to model the temporal dynamics of neuronal circuits. One framework is based on inferring latent neural dynamics via dynamic models. Within this framework, LFADS has shown great success in spiking neuronal datasets. This model consists of a sequential variational autoencoder that is tasked with reconstructing its input from a low-dimensional set of factors \cite{pandarinath2018inferring, zhu2022deep}. For continuous models, PLNDE has been successful in modeling  spiking neuronal dynamics via a Poisson neural differential equation model \cite{kim2021inferring}. Another approach has been to use encoding models to understand how neurons represent sensory inputs \cite{sinz2018stimulus,walker2019inception,bashiri2021flow}. These models are trained to reconstruct neuronal activity based on inputs such as images or sound sequences. This approach has been applied to spiking and 2-photon calcium data. However, it has not been used for whole-brain 1-photon calcium dynamics. Another category consists of goal-driven models, which are models trained to perform tasks that require human-like cognition in order to produce outputs that are correlated to neuronal brain dynamics \cite{yamins2014performance,yamins2016using,tang2018recurrent,cadena2019deep,li2022robust}. While such models have been widely employed to predict neuronal activity, they require complex experimental validations to infer meaningfulness.

\begin{figure*}[t]
\centering
  \includegraphics[width=0.95\textwidth]{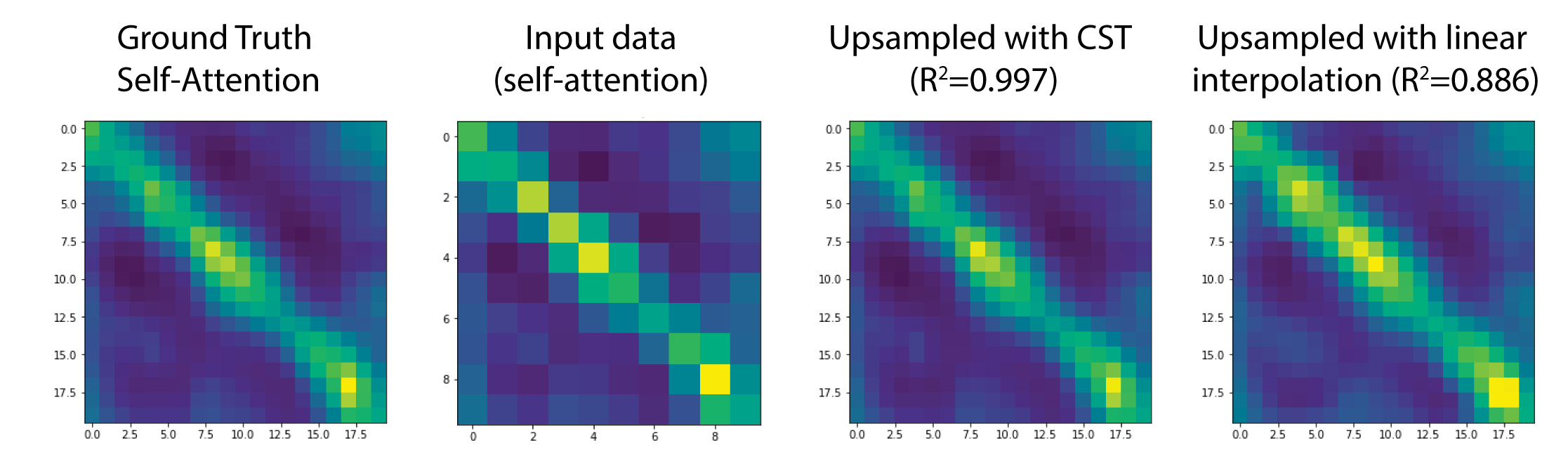}
  \caption{CST can accurately up-sample self-attention weights. Shown are, from left to right, attention maps for: ground truth data, down-sampled input data, up-sampling via CST, and up-sampling via linear interpolation (as performed in \citet{caron2021emerging}). We observe that CST provides up-sampled self-attention weights that more closely match the ground truth ($P<0.0001$) compared to linear interpolation (Figure \ref{fig:CST_SelfAttention_Upsampling_Error}). More examples are shown in Figure \ref{fig:CST_SelfAttention_Examples}.}
  \label{fig:CST_SelfAttention_UpSample}
\end{figure*}

\section{Method}
\label{sec:Method}

One of the essential components of the Transformer model is the positional encoding function, which specifies the order of the elements (or ‘tokens’) in the sequence and combines it with the encoded representation of the tokens. This is a successful approach for NLP and computer vision tasks, but too restrictive for datasets that are intrinsically continuous such as brain activity. Thus, we redesigned the Transformer to work more appropriately on the continuous domain. These modifications result in a new framework, called Continuous Spatiotemporal Transformer (CST\footnote{\href{https://github.com/vandijklab/CST}{https://github.com/vandijklab/CST}}) (Figure  \ref{fig:method}). During training, the model receives a mix of real (i.e., sampled data) and randomly sampled in-between-data (``dummy'') coordinates. The dummy points are initialized via a linear interpolation fitted on the sampled data points and evaluated at the dummy coordinates (Figure  \ref{fig:method}A). Next, this sequence of points is augmented via the addition of Gaussian noise (Figure  \ref{fig:method}B). Each point of the sequence is treated as a token, which is encoded via a linear encoder to a latent space and then fed into the Multi-Head Attention module (Figure  \ref{fig:method}C), which computes the self-attention between the tokens of the sequence \cite{vaswani2017attention}. We make use of the linear attention from \citet{xiong2021nystromformer}  which has $O(n)$ complexity, resulting in a linear increase of computational cost with the number of added points (Figure \ref{fig:computational_cost}). Finally, a linear decoder projects the tokens of the sequence from latent space to data space, resulting in the model's prediction. The model is optimized to minimize a Sobolev loss where the $p$-norm is computed between output and target data, while simultaneously minimizing the $p$-norm of higher derivatives. This prevents the formation of cusp points and other singularities in the interpolation output. During inference, no augmentation is performed and the model can be evaluated at any desired continuous coordinates within the data domain.

To better elucidate the aforementioned Sobolev loss optimization, first recall that a function $f\in C([a,b])$ is said to be weakly differentiable if there exists an integrable function $f'$ such that $\int_a^bf\phi' = -\int_a^b f'\phi$ for all differentiable functions $\phi\in C^1([a,b])$. Note that if a function is differentiable in the usual sense, then its weak derivative is easily seen to coincide with the notion of weak derivative (see e.g. \citet{Brezis}). For higher dimensional spaces a similar definition can be introduced as well. 
Then, the Sobolev space $W^{k,p}$ is inductively defined as the space of (weakly) $(k-1)$-differentiable functions $f$ with $f'\in W^{k-1,p}$, and with norm given by
\begin{equation}
||f||_{W^{k,p}}^p = ||f||_p^p + \sum_{q=1}^k ||D^qf||_p^p,
\end{equation}
where $D$ denotes the differential operator $D^q(f) := \frac{\partial^q f(x)}{\partial x^q}$.
As base of the induction definition, the case $k=1$ is defined as the space of weakly differentiable functions and equipped with norm given by
\begin{equation}
||f||_{W^{1,p}}^p = ||f||_p^p + ||f'||_p^p,
\end{equation}
where $f'$ indicates the weak derivative. When the domain space $\Omega\in \mathbb R^n$ is higher dimensional, the definition is the same as above, but we take into account all the partial derivatives indexed by multi-indices $\mathbf q$. See Appendix~\ref{sec:Sobolev}. 

Our optimization is performed in the Sobolev space, where we minimize the loss $\mathcal L$ defined as
\begin{equation}
\mathcal L(\mathbf y,\mathcal D)^p = ||\mathbf y_{\mathcal D}||_p^p + \mu*\sum_{|\mathbf q|=1}^k||D^{\mathbf q}(\mathbf y)||_p^p,
\end{equation}
where $\mathbf y$ is the output of the model, $\mathcal D$ indicates the data, $\mathbf y_{\mathcal D}$ is the function obtained as $\mathbf y_{\mathcal D} := y-{\mathcal D}$ and $\mu$ is a hyperparameter that regulates the contribution of higher derivatives to the optimization. Here we have used the multi-index notation $\mathbf q = (q_1, \ldots , q_r) \in \mathbb N^r$, and $D^{\mathbf q} := \frac{\partial^{q_1}\cdots\partial^{q_r}}{\partial x_1^{q_1}\cdots \partial x_r^{q_r}}$. For example, for $ \mathbf y(x_1,x_2,x_3)$, we have $D^{(1,0,3)}\mathbf y := \partial_1\partial^3_3 \mathbf y$, where we have used the notation $\partial_j^k := \frac{\partial^k}{\partial x_j^k}$. This parameter determines the wanted trade-off between accuracy for the model to fit the data, and the bound on the derivatives. In addition, $k$ and $p$ are also hyperparameters that define the Sobolev space in which the optimization is performed.
Observe that while the zero term takes into account the data, the higher degree terms do not refer to the data contrary to other approaches such as \cite{czarnecki2017sobolev}. This allows us to sample arbitrarily many points from the domain for the evaluation of the derivatives, for which we do not have data points. 

While it is conceptually desirable to have a model whose output, and its derivatives, can be sampled continuously, we also demonstrate by means of experimentation (see Section~\ref{sec:Experiments} below) that simply interpolating the output of a model does not necessarily give good interpolation results. In fact, in the presence of noise or irregularly sampled data, interpolating the output of the model using traditional polynomial methods can be negatively affected by fluctuations that cause overshooting. Our approach shows that when CST outputs the interpolated points through evaluation of the model itself, a lower interpolation error is obtained. As a further conceptual gain in our approach with CST, we can upsample the attention weights of CST via evaluation at any arbitrary point within the domain. As the model is shown to accurately predict the interpolated points, this attention results in a meaningful upsampling. 

Because CST combines both content and positional information of the data points to make predictions, the training forces the in-between coordinates to carry meaningful information about the modeled data. This allows us to use discretely sampled data to make predictions while generalizing for any arbitrary time point. This is important for computing smooth transitions between data points, facilitating interpolations, and eliminating the dependence upon regularly sampled data.

\section{Experiments}\label{sec:Experiments}

To benchmark CST with respect to continuity and smoothness, we have considered several synthetic and real-world datasets for which we have evaluated the interpolation error. Our experiments consistently show that while all models can fit the given datasets, CST outperforms them in interpolation tasks for noisy and irregularly sampled data.  

\subsection{Synthetic 2D spirals dataset}
\label{sec:2d_spirals}
To clearly show-case the properties of CST in comparison to the conventional Transformer, we use both methods for modeling 2D spirals generated by integral equations. This data consists of 500 2D spirals of 100-time points each. The data was split into 70\% of the spirals for training and 30\% for validation. For training, 10 data points were sampled from each curve while the remaining points were reserved for the interpolation test. Details about the data generation are described in Appendix~\ref{data_generation} and an example of a curve from this dataset is shown in Figure \ref{fig:2d_IE_spiral}. 

To train the Transformer model, we used the training procedure used by the authors of BERT \citep{devlin2018bert}. At each training step, we randomly select 30\% of the points for masking. The selected points are either replaced by a constant (80\% of the time), replaced by another random point (10\% of the time), or not replaced at all (10\% of the time). The model is trained to predict the data point selected for masking. Both CST and the Transformer have 4 layers, 4 heads, and $d_{model}$=32 (see Table \ref{tab:list_of_parameters_2DCurves} for more details).

To inspect the models, we sampled 1000 new time coordinates within the time interval of the data. The results obtained for CST and the Transformer model are shown in Figure \ref{fig:2D_IE_curves_CST_vs_Transformers}. We show that a Transformer model trained with the framework used in language modeling results in a step-like output, which yields poor interpolation performance. On the other hand, CST provides an output that better represents the original data. To evaluate the performance of both models in learning the dynamics of the dataset, we use the trained models to interpolate for the unseen data coordinates and compute the mean of the error per interpolated point. We show that CST has a significantly lower interpolation error ($P<0.0001$, N=150 spirals of the validation dataset) than the Transformer (Figure \ref{fig:2D_IE_curves_CST_vs_Transformers}C and D, Table \ref{tab:interpolation_error}). To illustrate that the lower interpolation error achieved by CST is not simply due to the augmentation during training, we performed the same augmentation for the Transformer model. The results show that the data augmentation with Gaussian noise harms the Transformer and induces higher interpolation error (Figure \ref{fig:interpolation_error_2}). 




Next, we compare CST's interpolation performance to common interpolation methods, such as linear and cubic spline interpolations. While simply performing interpolation is not our primary goal, we show that CST is more robust to noise than commonly used interpolation methods. To test this, we perturb the spirals with Gaussian noise $\mathcal{N}(0,0.1)$. We show that CST has a significantly lower interpolation error ($P<0.0001$, N=150 spirals of the validation dataset, Figure \ref{fig:2d_spiral_CST_vs_InterpolationMethods}) than linear and cubic spline interpolation. Examples of outputs are shown in Figure \ref{fig:2d_spiral_CST_vs_LinInterp_noisyData}.

Next, we show that CST can up-sample self-attention weights better than up-sampling via interpolation.
To test this, we use a model trained with 10 real points and 10 randomly sampled dummy points. During inference, we provide the same 10 points used during training and extra 10 fixed-coordinate points with their real values, thus providing a ground truth self-attention between the input points, as illustrated in Figure \ref{fig:CST_SelfAttention_UpSample} for a given input. To up-sample self-attention with CST, we provide the 10 points used during training and the time coordinates of the extra 10 points. We use CST to obtain outputs for all 20 points with their respective self-attention as described in Sec. \ref{sec:Method}. The self-attention obtained is illustrated in Figure \ref{fig:CST_SelfAttention_UpSample} (upsampled) for the same curve as in Figure \ref{fig:CST_SelfAttention_UpSample}. Another commonly used approach to up-sample attention is the use of interpolation methods \citep{caron2021emerging}. Here we use linear interpolation to up-sample the self-attention weights of the model's output for the 10 training points to 20 points. The result is shown in Figure \ref{fig:CST_SelfAttention_UpSample}. More examples are shown in Figure \ref{fig:CST_SelfAttention_Examples}. We evaluate the up-sampling performance of both approaches in terms of the attention error for the time coordinates not used during training. Figure \ref{fig:CST_SelfAttention_Upsampling_Error} shows the error distribution for CST and the linear interpolation in the up-sampling task for the validation curves. We observe that CST significantly ($P<0.0001$) outperforms the linear interpolation in terms of lower approximation error of self-attention up-sampling.

\subsection{Modeling dynamics in a video dataset}
\label{sec:benchmark_video}

Video recordings are a common type of data that benefits from continuous modeling. Although frames are discrete in time, they represent samples of a continuous process, and therefore, the dynamics in videos is conveniently modeled as such. In this section, we use CST to model dynamics in the KITTI video dataset \cite{geiger2013vision}. This dataset consists of recordings captured by a vehicle moving in the city of Karlsruhe. We utilized the version of the dataset presented in PredNet \cite{lotter2016deep} and split the dataset into 70\% for training and  30\% for validation. We modified the task to make it a video inpainting task by extracting 10 frames of the video sequence and adding gaussian noise ($\mathcal{N}(0,\sigma=0.5)$) to 40-60$\%$ of each frame, this noise perturbs the information in the frames (see Figure~\ref{fig:example_visualization_video_cst} for an example of a video sequence). Then, we trained the model to reconstruct the uncorrupted sequence based on the masked input.

We compared CST to other neural-networks-based models that are able to model spatiotemporal dynamics: ConvGRU \cite{ballas2015delving}, 3D-ViT and ViViT \citep{arnab2021vivit} (see Table \ref{tab:list_of_parameters_KITTI} for architecture details). ConvGRU was trained to recursively predict the frames from a single frame as input for every timepoint. 3D-ViT is a model based on the transformer architecture, and has 3-dimensional tokens for a 3d-tensor input. The ViViT model was trained following their factorized-encoder approach, wherein space and time are modeled by two separate Transformers. All models were trained on an RTX 3090 NVIDIA GPU for up to 150 epochs or until convergence. In Table \ref{tab:kitti_video} the validation mean squared error of the models trained on the video inpainting task is reported. We can observe that CST has a lower validation mean squared error compared to all other models. Furthermore, the reconstructed frames generated by CST have a lot more high-frequency similarities to the initial frame compared to other models (see examples in Figure \ref{fig:example_visualization_video_cst}).

\begin{table}[t]
\caption{Mean Squared Error on Video Inpainting Task for KITTI Dataset.}
\label{tab:kitti_video}
\vskip 0.15in
\begin{center}
\begin{small}
\begin{sc}
\begin{tabular}{lccr}
\toprule
  & Mean Squared Error \\
\midrule
ConvGRU  & 0.363 \\
ViViT & 0.3651 \\
3D-ViT & 0.2505 \\
\textbf{CST} & \textbf{0.1138}  \\
\bottomrule
\end{tabular}
\end{sc}
\end{small}
\end{center}
\vskip -0.1in
\end{table}

\subsection{Navier-Stokes equations}\label{sec:NS_experiment}

We consider a $(2+1)D$ PDE system, namely the Navier-Stokes equation \cite{chorin1968numerical,fefferman2000existence}, to evaluate the capability of CST to continuously model dynamical systems. The dataset consists of $5K$ instances of numerical solutions of the Navier-Stokes equation with random initial conditions. Further details on the dataset can be found in Appendix~\ref{sec:NS_dataset}. We trained CST on $1K$ dynamics and then tested the model on $300$ unseen noisy dynamics. Training is performed on $10$ time points of the dynamics, while testing is performed on a time sequence that includes $10$ additional time points that were unseen during training. Therefore, this is both an extrapolation task (with respect to the new initial condition of the dynamics), and an interpolation task (with respect to the unseen time points).

We compare CST with a Transformer model whose output is interpolated to obtain the predictions at data points between the given frames, and FNO2D and FNO3D \cite{li2020fourier}. We see that interpolation methods applied to the output of the transformers do not perform as well as CST, since they are negatively affected by noisy data. Moreover, we observe that while FNO3D is known to have achieved excellent results in interpolation tasks, the presence of irregularly sampled time points and noise greatly decreases the interpolation capabilities of the model, resulting in poor interpolation. We were unable to obtain good interpolations for FNO3D, despite properly fitting the data during training. 
The results of this experiment are shown in Table \ref{tab:NS_results}, and a list of parameters is given in Table \ref{tab:list_of_parameters_NS}.
Overall, this experiment shows that CST is able to learn continuous dynamics and that this model is a powerful tool when operating on noisy and sparsely sampled data.

\begin{table}[t]
\caption{Results for the interpolation task on the Navier-Stokes dataset reported as (mean $\pm$ std). The models were trained using $10$ time points per curve, and during inference, $20$ points randomly selected from the dynamics were predicted for unseen curves (i.e. new initial conditions).}
\label{tab:NS_results}
\vskip 0.15in
\begin{center}
\begin{small}
\begin{sc}
\begin{tabular}{lccr}
\toprule
  & $MSE$ \\
\midrule
Linear  & $(5.12 \pm 0.27)\times 10^{-3}$ \\ 
Spline & ($5.54  \pm 0.31)\times 10^{-3}$ \\ 
FNO3D &   ($1.38 \pm 0.08)\times 10^{0}$\\
FNO2D &  ($1.88 \pm 0.11\times 10^{-2}$\\
\textbf{CST } & \textbf{(4.88 $\pm$ 0.21)$\times \mathbf{10^{-3}}$} \\
\bottomrule
\end{tabular}
\end{sc}
\end{small}
\end{center}
\vskip -0.1in
\end{table}

\subsection{Learning brain dynamics from calcium imaging data}
\label{sec:CaImg}
Understanding how brain activity dynamics relate to cognition is a major open question in neuroscience \citep{macdowell2020low,cardin2020mesoscopic, vyas2020computation}. Since CST is able to model complex non-local continuous spatiotemporal dynamics from data, we use CST to learn brain dynamics from widefield calcium imaging in freely behaving mice. 

Widefield calcium imaging measures neuronal activity by measuring the amount of calcium influx in the cells \citep{chen2013ultrasensitive,cardin2020mesoscopic}. We use the data from \citet{lohani2020dual} in which the mice are presented with visual stimuli of varying contrasts.

The widefield imaging generates videos of shape $x \in \mathbb{R}^{H \times W \times T}$, where $(H,W)$ represents the spatial resolution of a frame and $T$ is the length of the video. To model brain dynamics, we want to account for how the different regions of the brain influence each other over time. To model the interaction between regions in space and time, we split the images into patches $x_P \in \mathbb{R}^{N \times P_0 \times P_1 \times T }$, where $N=\left(HW\right)/P_0 P_1$ is the total number of patches per frame for a patch of size $P_0 \times P_1$. A similar approach is used in \cite{dosovitskiy2020image} and \cite{he2022masked}. 

\begin{figure}[t]
\centering
  \includegraphics[width=\columnwidth]{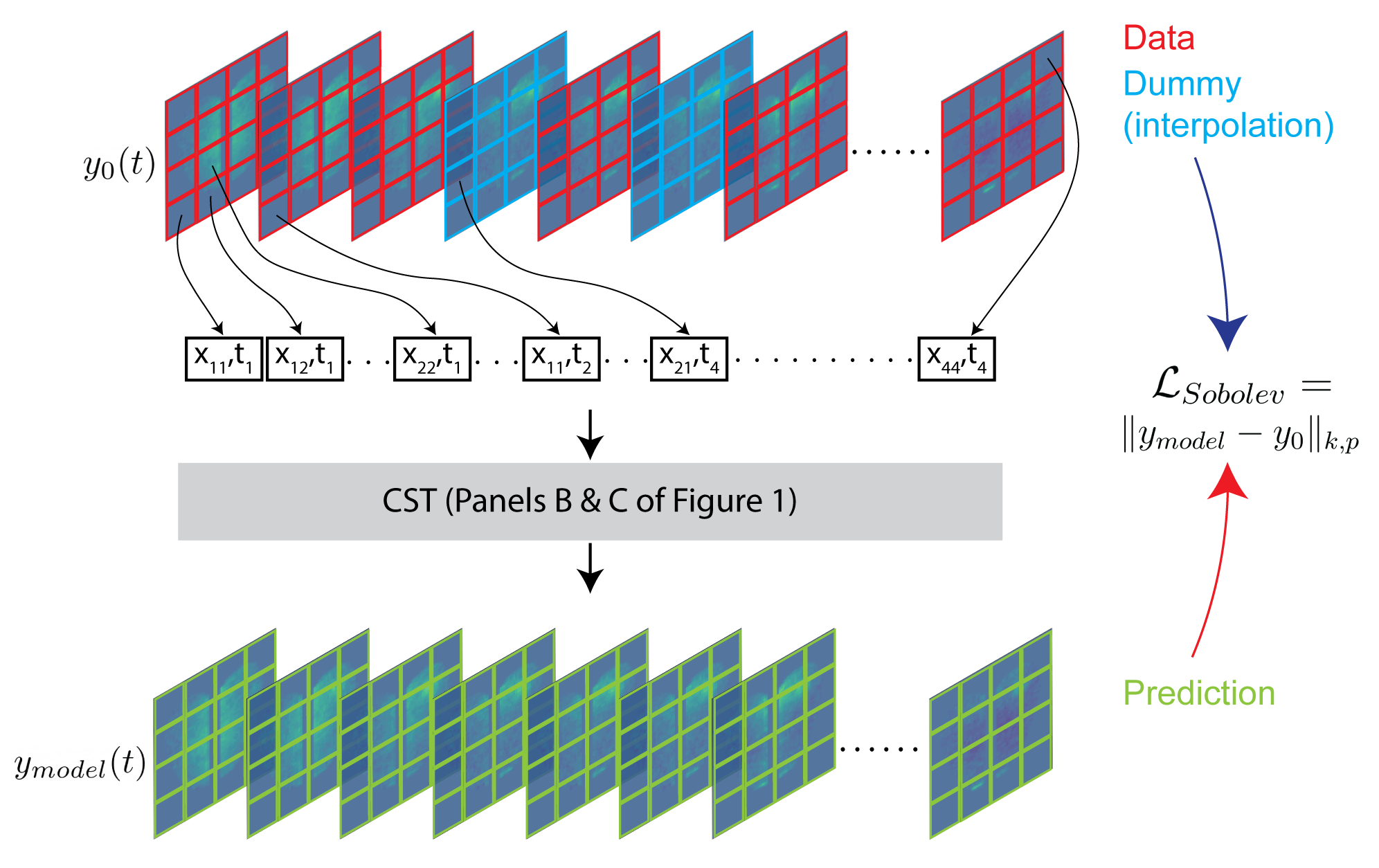}
  \caption{Diagram of CST spatiotemporal encoding for calcium imaging recordings. Each frame (time point) of the recording is partitioned into patches. Dummy frames are inserted using the procedure described in Sec. \ref{sec:CaImg}. Each patch is treated as a token of the sequence and is combined with its positional information (i.e., position in space and time). These tokens are encoded by CST as described in Figure \ref{fig:method}B and C. Loss of the model output w.r.t. the input data is computed in Sobolev space.}
  \label{fig:CST_Patches_Method}
\end{figure}

\begin{figure*}[t]
\centering
  \includegraphics[width=\textwidth]{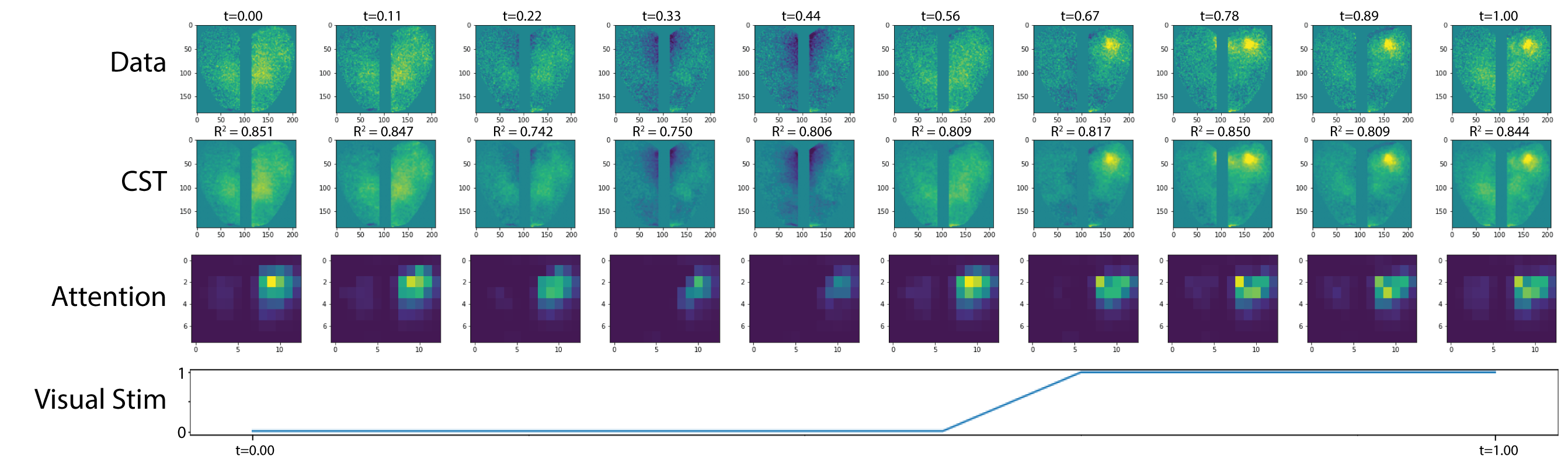}
  \caption{Modeling widefield calcium imaging with CST. The first row illustrates the input data. The second row shows CST's prediction. The coefficient of determination ($R^2$) indicates the quality of the prediction. The third row shows the corresponding attention weights obtained between the last frame of the sequence and the different regions of the previous frames. The fourth row shows changes in the visual stimuli, which is used to illustrate changes in neural activity and attention caused by changes in stimuli. See Figure \ref{fig:CalImg_Attention} for more examples.}
  \label{fig:CalImg_Attention2}
\end{figure*}

Next, the patches are perturbed and treated as tokens, such as described in Section \ref{sec:Method} and illustrated in Figure \ref{fig:CST_Patches_Method}. For this experiment, the frames of size $H=184$ and $W=208$ were divided into patches of size $P_0=23$ and $P_1=16$, making a total of $N=104$ patches per frame. The video is presented to the model as multiple segments of 10 frames. The recording is split in 70/30 for training and validation. During training, the patches are perturbed with noise $\mathcal{N}(0,\sigma=0.1)$. Figure \ref{fig:CalImg_Attention2} shows CST's performance in modeling a segment from the validation, and the corresponding $R^2$ coefficients. In addition to neural activity prediction, we also inspect the learned attention weights of the last frame of the sequence. Among the learned attention weights, some show patterns that match specific locations of the brain, such as the visual cortex (Figure \ref{fig:CalImg_Attention2}).

Next, we test to which extent the learned attention weights encode information about the visual stimulus in comparison to the raw data (baseline). To test this, we used PCA to reduce the dimensionality of the attention distribution from its original flattened shape (tokens $\times$ tokens) to 10 principal components and used a regression model to predict the contrast of the visual stimulus associated with the last frame of the sequence. We used 500 segments from the validation set, from which 70\% were used to train a KNN regressor (k=3) and the remaining 30\% were used for testing. The process of partitioning the data and fitting the regressor was repeated 10 times. To evaluate the influence of the number of parameters, we report CST-base (Transformer with 12 layers, 12 heads) and CST-small (6 layers, 6 heads) (see Table \ref{tab:list_of_parameters_CaImg} for more details). The mean square error (MSE) and the coefficient of determination ($R^2$) between the contrast of the visual stimulus presented and the prediction are shown in Table \ref{tab:regression_visual_stim}. Both CST-base and CST-small showed significantly higher $R^2$ and significantly lower MSE than the baseline.

We compare CST to other neural network-based models that are capable of learning a latent representation of the dynamics. The models and their performances are listed in Table \ref{tab:regression_visual_stim} (see Table \ref{tab:list_of_parameters_CaImg} for architecture details). All the architectures were decided in order to have models roughly with the same number of parameters and allow them to be trained and converge within 2 days on an RTX 3090 NVIDIA GPU. The LatentODE \citep{rubanova2019latent} was trained with a reversed RNN-encoder from the last to the first frame and then predicted the whole sequence from the encoded latent space. The LSTM \citep{hochreiter1997long} was trained to recursively predict the frames from the sequence with windows of 5 frames as input. LFADS was trained to reconstruct 10 frames of the sequence through a bottleneck layer of 100 dimensions (factors). The results shown in Table \ref{tab:regression_visual_stim} reflect the poor capacity of these models in learning the dynamics of brain activity, which spans from the low information contained in their latent spaces. We hypothesize that the non-local spatial dynamics induced by high contrast stimuli are responsible for the decreased performance of other models compared to CST. 

Next, we trained two models that account for spatially structured data: ConvLSTM \citep{shi2015convolutional} and ViViT. The ConvLSTM was trained similarly to the setup described for the LSTM. Despite ViViT being significantly larger than CST (Table \ref{tab:list_of_parameters_CaImg}), we observe that both CST-base and CST-small present more meaningful latent spaces with respect to encoding relevant information about the dynamics that describe the response to visual stimuli (Table \ref{tab:regression_visual_stim}). 


\begin{table}[t]
\caption{Performance of a KNN Regressor in regressing the contrast of visual stimuli from the learned latent representation. Results presented as (mean $\pm$ std, N=10).}
\label{tab:regression_visual_stim}
\vskip 0.15in
\begin{center}
\begin{small}
\begin{sc}
\begin{tabular}{lcccr}
\toprule
  & $MSE$ & $R^2$\\
\midrule
Data (baseline)  & 0.112 $\pm$ 0.020 & 0.595 $\pm$ 0.053 \\
LatentODE & 0.361 $\pm$ 0.051 & -0.316 $\pm$ 0.181 \\
LSTM  & 0.268 $\pm$ 0.029 & 0.024 $\pm$ 0.093 \\
ConvLSTM  & 0.180 $\pm$ 0.025 & 0.342 $\pm$ 0.103\\
LFADS & 0.112 $\pm$ 0.025 & 0.609 $\pm$ 0.069\\
ViViT & 0.101 $\pm$ 0.025 & 0.637 $\pm$ 0.045\\
\textbf{CST-small} & \textbf{0.090 $\pm$ 0.025} & \textbf{0.676 $\pm$ 0.064}\\
\textbf{CST-base} & \textbf{0.063 $\pm$ 0.019} & \textbf{0.774 $\pm$ 0.057}\\
\bottomrule
\end{tabular}
\end{sc}
\end{small}
\end{center}
\vskip -0.1in
\end{table}

\section{Conclusions}
We have presented the Continuous Spatiotemporal Transformer (CST), a new framework for applying Transformers to continuous systems. We have demonstrated CST's ability to learn the dynamics of several systems and the benefits that stem from a continuous representation. CST shows better interpolation performance compared to other methods and is capable of effectively up-sampling self-attention weights. Finally, we demonstrated CST on modeling brain dynamics from calcium imaging recordings. We showed that the latent space learned by CST is more informative of behavioral relevant variables, such as visual stimulus contrast, compared to other models. Furthermore, the self-attention weights learned by CST provide a biologically meaningful representation of the underlying brain dynamics. We anticipate that CST can be used to model spatiotemporal dynamical systems from other domains. 

\nocite{langley00}

\bibliography{references_icml.bib}
\bibliographystyle{icml2023}

\newpage
\appendix
\onecolumn

\section{Sobolev spaces}\label{sec:Sobolev}

In this appendix we provide definitions and some standard results on Sobolev spaces. The main references for this appendix are \cite{Brezis,atkinson2005theoretical}. Throughout this section we use the same multi-index notation of Section~\ref{sec:background}, where $\mathbf q = (q_1, \ldots, q_d)\in \mathbb N^d$ represents a vector consisting of non-negative integers (of some length $d$), and we set $D^{\mathbf q}:= \frac{\partial^{q_1}\cdots\partial^{q_d}}{\partial x_1^{q_1}\cdots \partial x_r^{q_d}}$ to indicate the partial differential operator corresponding to $\mathbf q$. Moreover, we set $|\mathbf q| = \sum_i q_i$. 

For completeness, we recall now the definition of weak derivative, which is usually used to define the Sobolev space, even though our functions are all differentiable, and the notion of weak derivative simply coincides with the regular differentiation. 

\begin{definition}
{\rm 
Let $\Omega \subset \mathbb R^d$ be a nonempty open set, and let $f,g$ be locally integrable functions. Then we say that $g$ is the weak $\mathbf q^{\rm th}$ derivative of $f$ if it holds
\begin{equation}\label{eqn:weak_der}
\int_\Omega f D^{\mathbf q} \phi d\mathbf x = (-1)^{|\mathbf q|}\int_\Omega g \phi d\mathbf x,
\end{equation}
for all smooth $\phi$ such that $\phi|_{\partial \bar\Omega} = 0$.
}
\end{definition}

\begin{remark}
{\rm 
Observe that when $f$ is differentiable (with respect to the $D^{\mathbf q}$ operator), then $D^{\mathbf q}f$ satisfies Equation~\ref{eqn:weak_der} by an integration by parts argument. As a consequence, the weak derivative coincides with the regular derivative almost everywhere (i.e. up to a subset of measure zero). 
}
\end{remark}

While the notion of Sobolev space exists in very general contexts, we have considered in this article only the spaces of integer order, and we therefore hereby provide only the definition of these objects.

\begin{definition}
{\rm 
Let $k$ be non-negative integers, let $p \in (\mathbb N-\{0\})\cup \{\infty\}$, and let $\Omega$ be a domain in $\mathbb R^d$. The Sobolev space $W^{k,p}(\Omega)$ is defined to be the space of $p-$integrable functions (i.e. functions in $L^p(\Omega)$) such that for all multi-index $\mathbf q$ such that $|\mathbf q| \leq k$, the $\mathbf q^{\rm th}$ derivative exists and it is $p$-integrable. The norm in $W^{k,p}(\Omega)$ is defined to be
\begin{equation}
||f||_{W^{k,p}(\Omega)} = \begin{cases}
[\sum_{|\mathbf q|\leq k} ||D^{\mathbf q}f||_{L^p(\Omega)}]^{1/p}\ \ p\neq \infty \\
\max_{|\mathbf q|\leq k} ||D^{\mathbf q}f||_{L^p(\Omega)}\ \ p=\infty.
\end{cases}
\end{equation}
}
\end{definition}

We recall this very important result.

\begin{theorem}
The Sobolev space $W^{k,p}(\Omega)$ is a Banach space for all $k,p$. Moreover, the space $H^k(\Omega) := W^{k,2}(\Omega)$ is a Hilbert space.
\end{theorem}

We illustrate the use of Sobolev spaces in the study of differential equations by considering a standard example of variational formulation of boundary value problem.  

We consider the Poisson boundary value problem 
\begin{equation}
\begin{cases}
    -\Delta u = f \ {\rm in}\ \Omega \\
    u_{|\partial \bar\Omega} = 0,
\end{cases}
\end{equation}
where $f$ is a smooth function. 
A solution of the Poisson boundary value problem is a smooth function ($C^2$-class) which is continuous on the boundary $\Gamma := \partial\bar\Omega$. The idea now is that if such a solution $u$ exists, then for all smooth functions $v$ vanishing on $\Gamma$ it holds that
\begin{equation}
    - \int_\Omega \Delta u d\mathbf x = \int_\Omega f v d\mathbf x.
\end{equation}
Integrating by parts, and making use of the fact that $v$ vanishes on the boundary $\Gamma$, one obtains the equality
\begin{equation}\label{eqn:weak}
    \int_\Omega \nabla u \cdot \nabla v d\mathbf x = \int_\Omega f v d\mathbf x,
\end{equation}
which is the {\it weak formulation} of the Poisson boundary value problem, and makes sense when seeking solutions $u$ in the Sobolev space $H^1_0(\Omega) \subset H^1(\Omega)$ consisting of the functions in $H^1(\Omega)$ which are trivial on the boundary $\Gamma$, and it should be solved for all $v\in H^1_0(\Omega)$. When solving Equation~\ref{eqn:weak}, the Poisson equation is satisfied only up to a subset of $\Omega$ of measure zero, due to the presence of the integral. However, if the solution is regular enough, equality holds strictly in the classical sense, and we obtain a solution of the original problem. The convenience of this formulation is that we can now apply strong results in Banach/Hilbert spaces to show that a unique solution exists.

\section{Sobolev loss implementation}

We give here a pseudocode for the implementation of the Sobolev loss employed in this article to train the CST model. 

\begin{algorithm}
    \caption{Implementation of the Sobolev loss. The model takes an input on the real points, interpolates on dummy points and produces an output. The values corresponding to real points are used to compare with the ground truth, and the values on the dummy points are used to compute higher derivatives. The norms corresponding to all the elements, both real and dummy points, are summed. A corresponding implementation that includes spatial derivatives can be obtained by a straightforward generalization of this algorithm.}
    \label{algo:Sobolev}
    \begin{algorithmic}
    \REQUIRE{$\{\mathbf y(t_i)\}$, $\{\bar t_j\}$} \ \ \COMMENT{Initialization at given time points and dummy points}
    \ENSURE{$\{\mathbf y(\hat t_k)\}$}\ \  \COMMENT{Output values of CST at original and dummy points $\hat t_k$}
    \STATE{$\mathbf y_0 = \mathbf y(\hat t_k)$, $L = \mathbf y(t_i) - \mathcal D$} \ \
    \COMMENT{Obtain $\mathbf y$ for all points (real and dummy) using interpolation, and set loss to be difference between observed points and data}
    \WHILE{$i<k$:}
        \STATE{$\mathbf y_{i+1} = \partial_{\hat t} T(\mathbf y_i)$}\ \ \COMMENT{Compute the derivatives of input (including dummy points) for model $T$}
        \STATE{$L \xleftarrow{+} ||\mathbf y_{i+1}||^p$}\ \ \COMMENT{Add $p$-norm of the $i^{\rm th}$ derivative to the loss}
    \ENDWHILE
    \STATE{$L_{out} = L^{1/p}$}
    \end{algorithmic}  
\end{algorithm}

\section{Artificial Dataset Generation}\label{data_generation}

\subsection{Integral Equation Spirals}

We defined an Integral Equation system for a 2D spiral. We then solve the system using an integral equation solver in Pytorch numerically, without the need of a neural network for the forward function. See c, d, k and f which defines the dynamical system in IESolver monoidal. See code below and example dynamics:

\begin{figure}[ht]
\centering
  \includegraphics[width=1.0\textwidth]{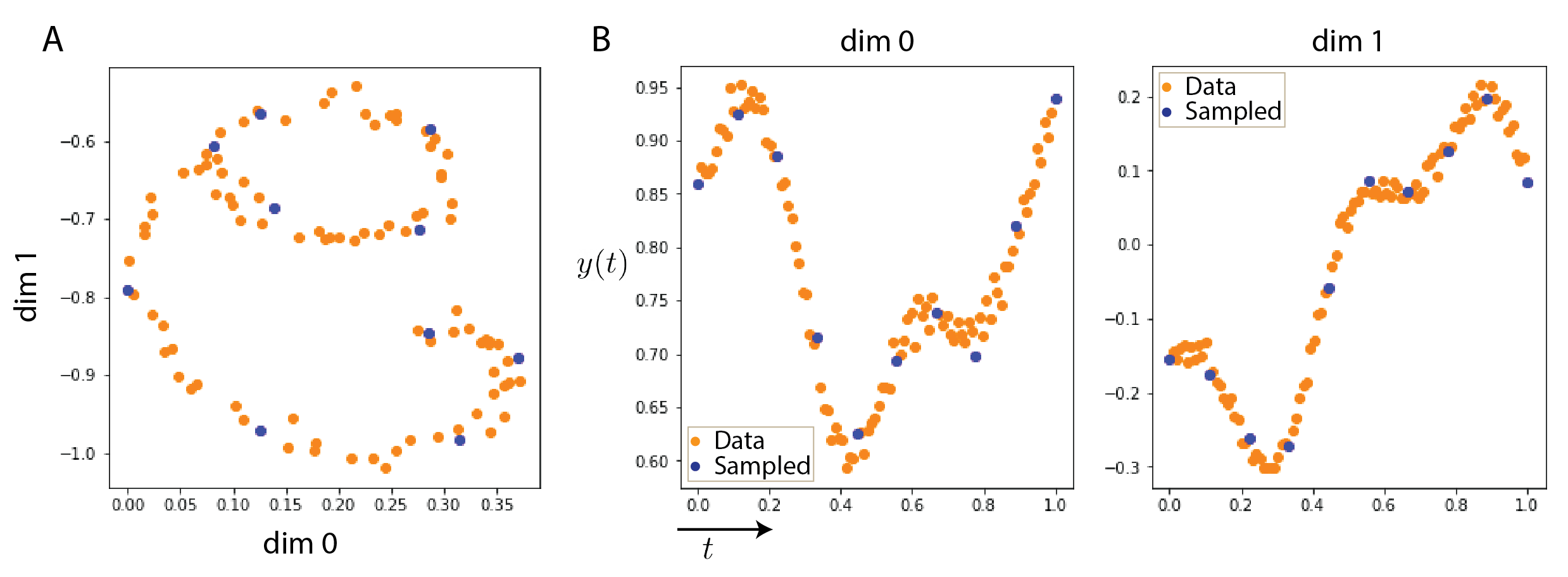}\caption{Example visualization of 2D Integral Equation Spiral}
  \label{fig:2d_IE_spiral}
\end{figure}

\begin{lstlisting}
z0 = torch.Tensor([[0.1, 0.7]]).to(device)

t_max = 1
t_min = 0
n_points = 100

index_np = np.arange(0, n_points, 1, dtype=int)
index_np = np.hstack([index_np[:, None]])
times_np = np.linspace(t_min, t_max, num=n_points)
times_np = np.hstack([times_np[:, None]])

###########################################################
times = torch.from_numpy(times_np[:, :, None]).to(z0)
times = times.flatten()
#times = times/t_max
###########################################################
start = time.time()
solver = IESolver_monoidal(x = times.to(device),

c = lambda x: torch.Tensor([torch.cos(torch.Tensor([x])),
torch.cos(torch.Tensor(x+np.pi))]).to(device),

d = lambda x,y: torch.Tensor([1]).to(device), 
k = lambda t,s: kernels.cos_kernel(2*np.pi*t,-2*np.pi*s), 
f = lambda y: torch.tanh(2*np.pi*y).to(device),
lower_bound = lambda x: torch.Tensor([t_min]).to(device),
upper_bound = lambda x: x,
max_iterations = 3,
integration_dim = 0,
mc_samplings = 10000)

Data = solver.solve()
Data = Data.unsqueeze(1)

\end{lstlisting}

\subsection{Navier-Stokes equation}\label{sec:NS_dataset}
The Navier-Stokes equations are partial differential equations that arise in fluid mechanics, where they are used to describe the motion of viscous fluids. They are derived from the conservation laws for Newtonian fluids subject to an external force with the addition of pressure and friction forces, where the unknown function indicates the velocity vector of the fluid \cite{chorin1968numerical,fefferman2000existence}. Their expression is given by the system

\begin{equation}
    \frac{\partial}{\partial t}u_i + \sum_j u_j\frac{\partial u_i}{\partial x_j} = \nu \Delta u_i - \frac{\partial p}{\partial x_i} + f_i(\bold{x},t)
\end{equation}
\begin{equation}
    {\rm div} u = \sum_i \frac{\partial u_i}{\partial x_i}
\end{equation}
where $\Delta$ is the Laplacian operator, $f$ is the external force, and $\bold u$ is the unknown velocity function. We experiment on the same data set for $\nu = 1e-3$ of \cite{li2020fourier}, which can be found in their GitHub page \footnote{https://github.com/zongyi-li/fourier\_neural\_operator/tree/master/data\_generation/navier\_stokes}. We use $4000$ instances for training and $1000$ for testing.

\section{Data collection}\label{sec:data_collection}

Here we report details about data collection and preprocessing.  

\subsection{Calcium imaging dataset}
\label{appendix:calcium_imaging}
C57BL/6J mice were kept on a 12h light/dark cycle, provided with food and water ad libitum, and housed individually following headpost implants. Imaging experiments were performed during the light phase of the cycle. For mesoscopic imaging, brain-wide expression of jRCaMP1b was achieved via postnatal sinus injection as described in  \cite{barson2020simultaneous,hamodi2020transverse}. 

Briefly, P0-P1 litters were removed from their home cage and placed on a heating pad. Pups were kept on ice for 5 min to induce anesthesia via hypothermia and then maintained on a metal plate surrounded by ice for the duration of the injection. Pups were injected bilaterally with 4 ul of AAV9-hsyn-NES-jRCaMP1b ($2.5\times10^{13}$ gc/ml, Addgene). Mice also received an injection of AAV9-hsyn-ACh3.0 to express the genetically encoded cholinergic sensor ACh3.0, (Jing et al., 2020, although these data were not used in the present study. Once the entire litter was injected, pups were returned to their home cage.

Surgical procedures were performed on sinus-injected animals once they reached adulthood ($>$P50). Mice were anesthetized using 1-2\% isoflurane and maintained at 37ºC for the duration of the surgery. For mesoscopic imaging, the skin and fascia above the skull were removed from the nasal bone to the posterior of the intraparietal bone and laterally between the temporal muscles. The surface of the skull was thoroughly cleaned with saline and the edges of the incision secured to the skull with Vetbond. A custom titanium headpost for head fixation was secured to the posterior of the nasal bone with transparent dental cement (Metabond, Parkell), and a thin layer of dental cement was applied to the entire dorsal surface of the skull. Next, a layer of cyanoacrylate (Maxi-Cure, Bob Smith Industries) was used to cover the skull and left to cure ~30 min at room temperature to provide a smooth surface for trans-cranial imaging.

Mesoscopic calcium imaging was performed using a Zeiss Axiozoom with a 1x, 0.25 NA objective with a 56 mm working distance (Zeiss). Epifluorescent excitation was provided by an LED bank (Spectra X Light Engine, Lumencor) using two output wavelengths: 395/25 (isosbestic for ACh3.0, Lohani et al., 2020) and 575/25nm (jRCaMP1b). Emitted light passed through a dual camera image splitter (TwinCam, Cairn Research) then through either a 525/50 (ACh3.0) or 630/75 (jRCaMP1b) emission filter (Chroma) before it reached two sCMOS cameras (Orca-Flash V3, Hamamatsu). Images were acquired at 512x512 resolution after 4x pixel binning. Each channel was acquired at 10 Hz with 20 ms exposure using HCImage software (Hamamatsu).

For visual stimulation, sinusoidal drifting gratings (2 Hz, 0.04 cycles/degree were generated using custom-written functions based on Psychtoolbox in Matlab and presented on an LCD monitor at a distance of 20 cm from the right eye. Stimuli were presented for 2 seconds with a 5 second inter-stimulus interval

Imaging frames were grouped by excitation wavelength (395nm, 470nm, and 575nm) and downsampled from 512$\times$512 to 256$\times$256 pixels. Detrending was applied using a low pass filter (N=100, $f_{cutoff}=$0.001Hz). Time traces were obtained using $(\Delta F/F)_i=(F_i-F_{(i,o)} )/F_{(i,o)}$ where $F_i$ is the fluorescence of pixel $i$ and $F_{(i,o)}$ is the corresponding low-pass filtered signal.

Hemodynamic artifacts were removed using a linear regression accounting for spatiotemporal dependencies between neighboring pixels.  We used the isosbestic excitation of ACh3.0 (395 nm) co-expressed in these mice as a means of measuring activity-independent fluctuations in fluorescence associated with hemodynamic signals.  Briefly, given two $p\times1$  random signals $y_1$ and $y_2$ corresponding to $\Delta F/F$ of $p$ pixels for two excitation wavelengths “green” and "UV", we consider the following linear model:

\begin{eqnarray}
	y_1=x+z+\eta,
\end{eqnarray}
\begin{eqnarray}
	y_2=Az+\xi,
\end{eqnarray}	

where x and z are mutually uncorrelated $p\times1$ random signals corresponding to $p$ pixels of the neuronal and hemodynamic signals, respectively. $\eta$ and $\xi$ are white Gaussian $p\times1$ noise signals and A is an unknown $p\times p$ real invertible matrix. We estimate the neuronal signal as the optimal linear estimator for $x$ (in the sense of Minimum Mean Squared Error):

\begin{eqnarray}
	\hat{x} &=& H\left(\begin{array}{c} y_1  \\ y_2  \end{array}\right),\\
	H &=& \sum_{xy}{\sum_{y}}^{-1}
\end{eqnarray}

where $y=\begin{pmatrix} y_1  \\ y_2  \end{pmatrix}$ is given by stacking $y_1$  on top of $y_2$,  $\sum_y=E[yy^T ]$ is the autocorrelation matrix of $y$ and $\sum_{xy}=E[xy^T ]$ is the cross-correlation matrix between $x$ and $y$. The matrix $\sum_y$ is estimated directly from the observations, and the matrix $\sum_{xy}$ is estimated by:

\begin{eqnarray}
	\sum_{xy}=\Biggl(\sum_{y_1}-\sigma_{\eta}^{2}I- \biggl(\sum_{y_1 y_2} {\Bigl(\sum_{y_2}-\sigma_{\xi}^{2}I \Bigl)}^{-1} {\sum_{y_2}}^{-1} {\sum_{y_1 y_2}}^T \biggl)^T&0\Bigg) 
\end{eqnarray}

where $\sigma_\eta^2$ and $\sigma_\xi^2$  are the noise variances of $\eta$ and $\xi$, respectively, and $I$ is the $p\times p$ identity matrix. The noise variances $\sigma_\eta^2$ and $\sigma_\xi^2$  are evaluated according to the median of the singular values of the corresponding correlation matrices  $\sum_{y_1}$and $\sum_{y_2}$.  This analysis is usually performed in patches where the size of the patch, $p$, is determined by the amount of time samples available and estimated parameters. In the present study, we used a patch size of $p=9$.   The final activity traces were obtained by z-scoring the corrected $\Delta F/F$ signals per pixel. The dimensionality of the resulting video is then reduced via PCA to 10 components, which represents $\approx 80\%$ of data variance.

\section{2D IE spirals}\label{sec:2D_IE_curves_moreExamples}
More examples of outputs obtained with different models for the 2D IE spirals.

\begin{figure}[ht]
\centering
  \includegraphics[width=1.0\textwidth]{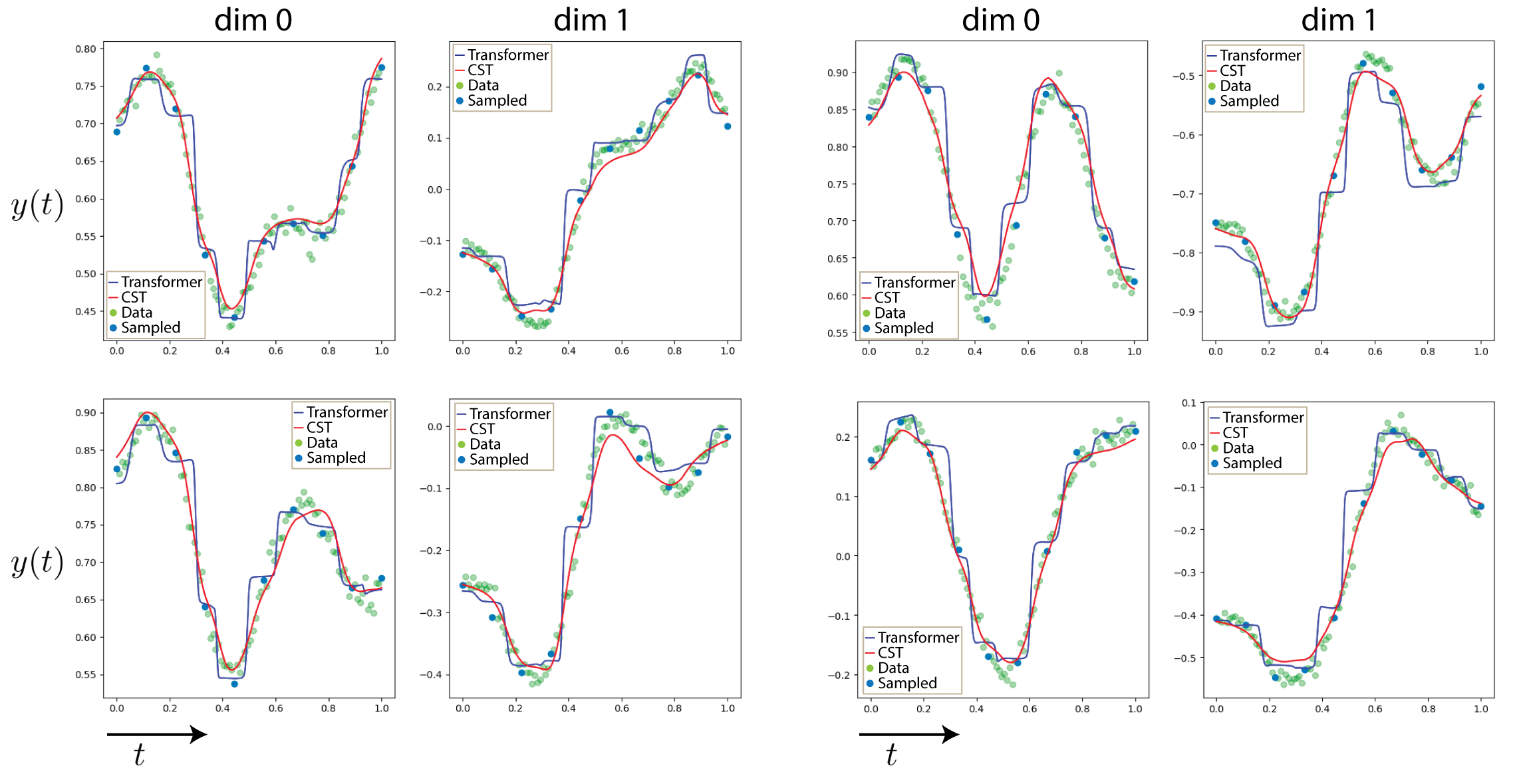}\caption{Examples of 2D IE curves modeled by the Transformer and CST. We observe that the Transformer presents a step-like output while CST results in smoother output and lower interpolation error (see Sec. \ref{sec:2d_spirals}).}\label{fig:2d_spiral_moreExamples}
\end{figure}

\begin{figure}[ht]
\centering
  \includegraphics[width=1.0\textwidth]{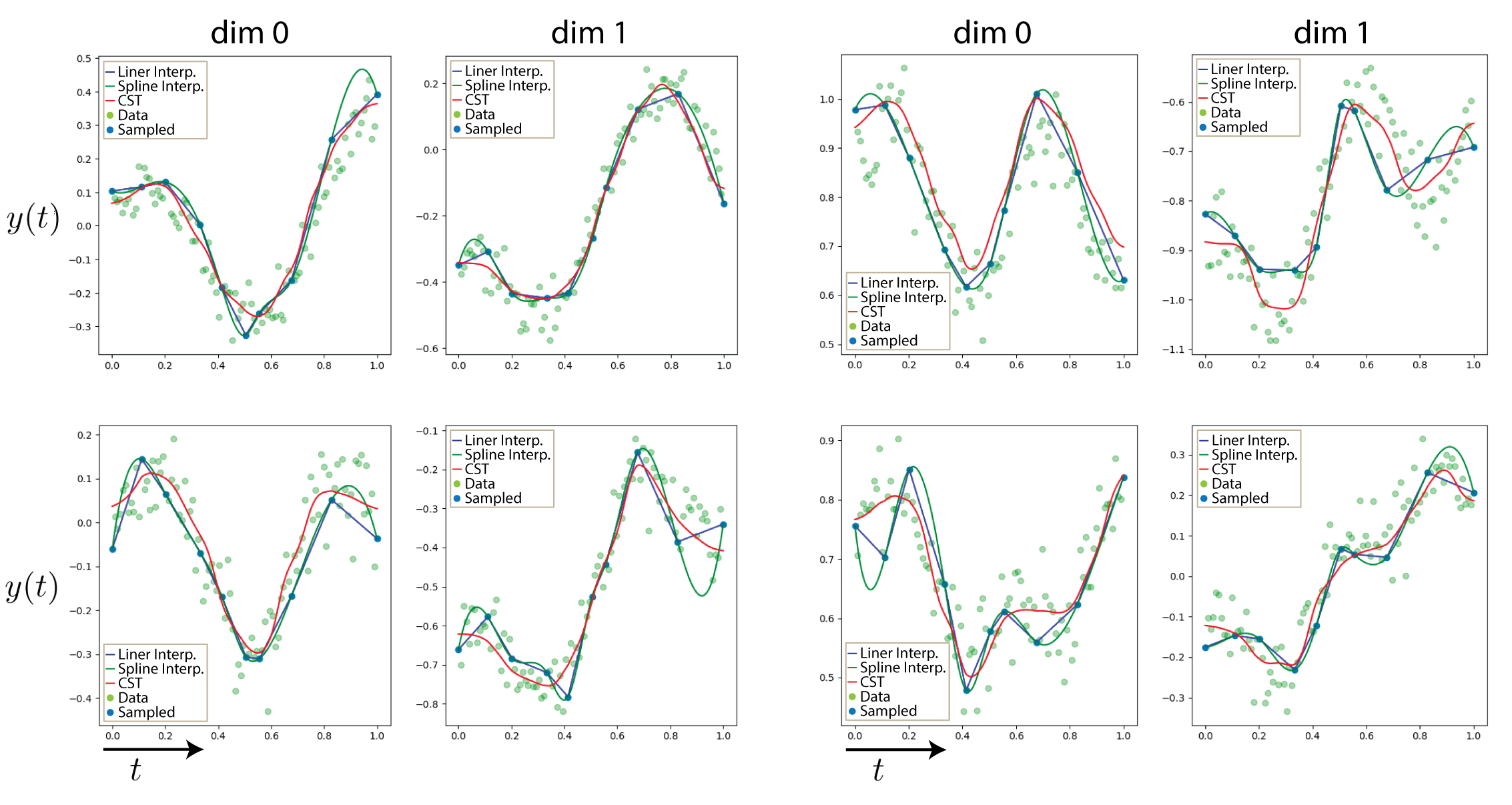}\caption{Examples of noisy 2D IE curves modeled by CST in comparison to Linear and Cubic Spline interpolation. We observe that interpolation methods are sensitive to noise than CST, which results in a lower interpolation error for CST (see Sec. \ref{sec:2d_spirals}).} \label{fig:2d_spiral_CST_vs_LinInterp_noisyData}
\end{figure}

\begin{figure}[ht]
\centering
  \includegraphics[width=0.5\textwidth]{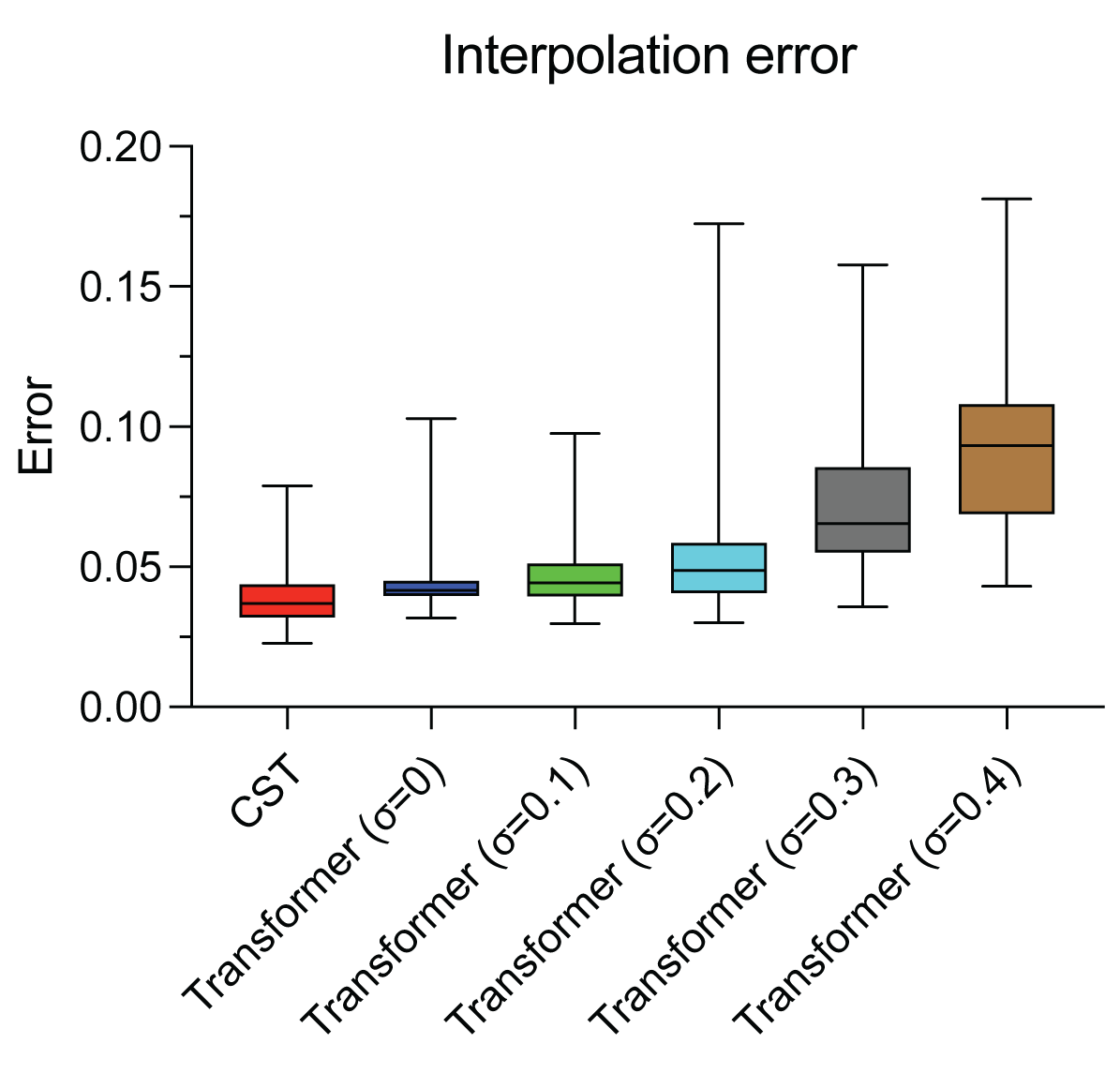}\caption{The interpolation error for CST and Transformer on the 2D IE curves. Here we include the error for the Transformer when trained with no data  augmentation via added noise ($\sigma=0$) or augmented with Gaussian noise $\mathcal{N}(0,\sigma)$  (see Sec. \ref{sec:2d_spirals}).}\label{fig:interpolation_error_2}
\end{figure}

\begin{figure}[ht]
\centering
  \includegraphics[width=0.6\textwidth]{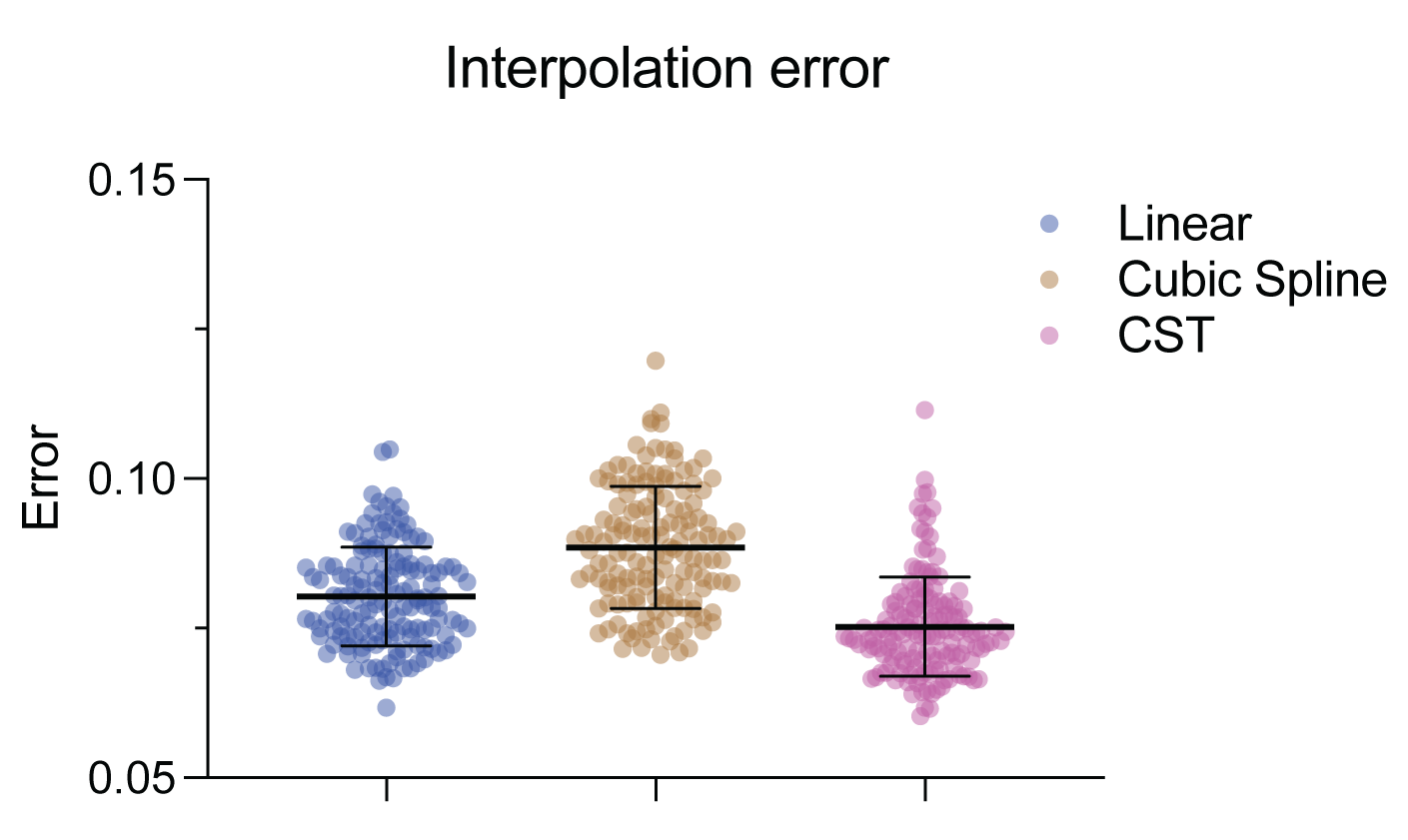}\caption{Distribution of interpolation error (L2-norm) for CST in comparison to linear and cubic spline interpolation (see Sec. \ref{sec:2d_spirals}).}\label{fig:2d_spiral_CST_vs_InterpolationMethods}
\end{figure}

\begin{figure}[t]
\centering
  \includegraphics[width=0.75\textwidth]{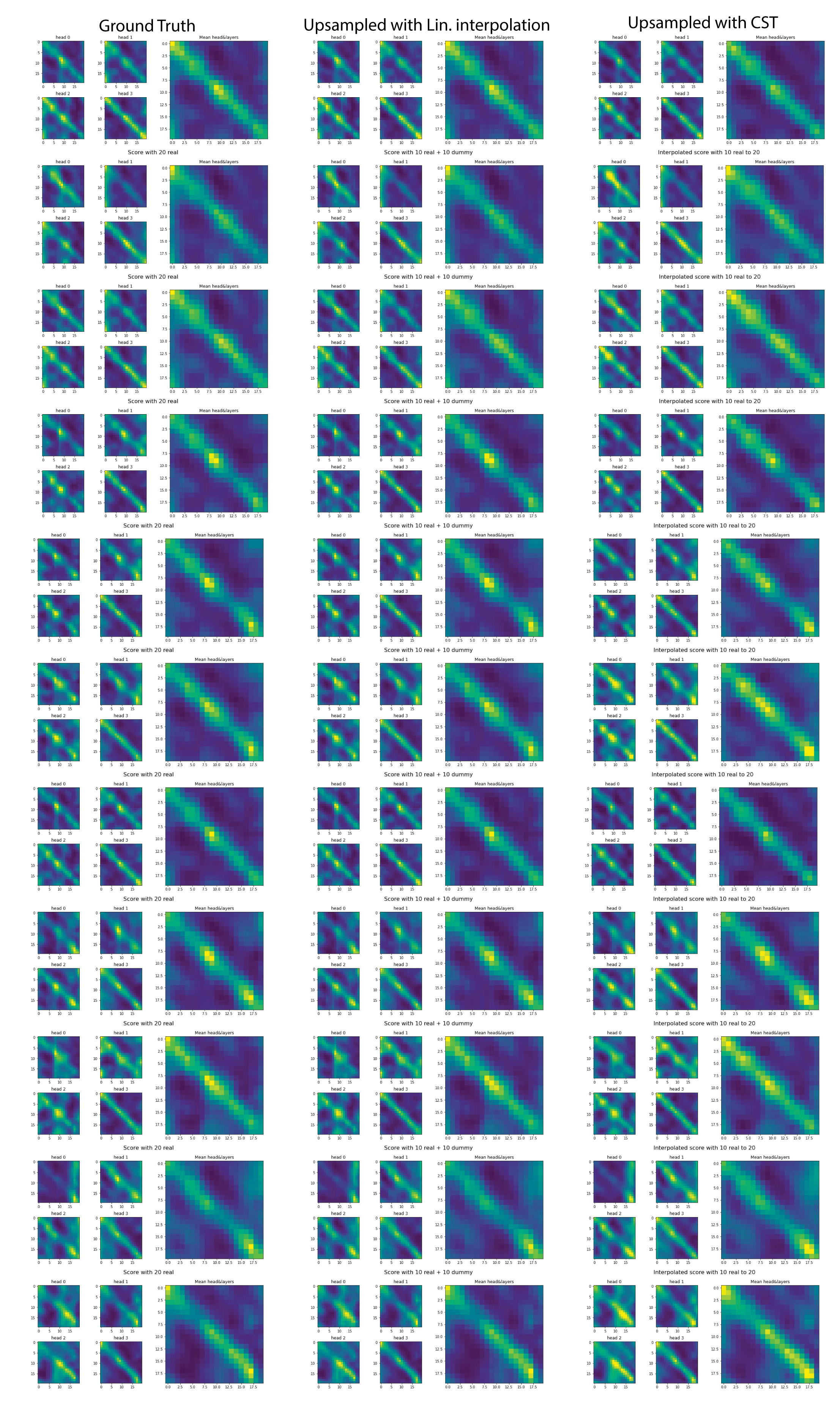}
  \caption{Examples of up-sampled attention with CST (middle column) vs. up-sampling via interpolation (right column) in comparison to the ground truth attention (left column) (see Sec. \ref{sec:2d_spirals}). }
  \label{fig:CST_SelfAttention_Examples}
\end{figure}

\begin{figure}[t]
\centering
  \includegraphics[width=0.4\textwidth]{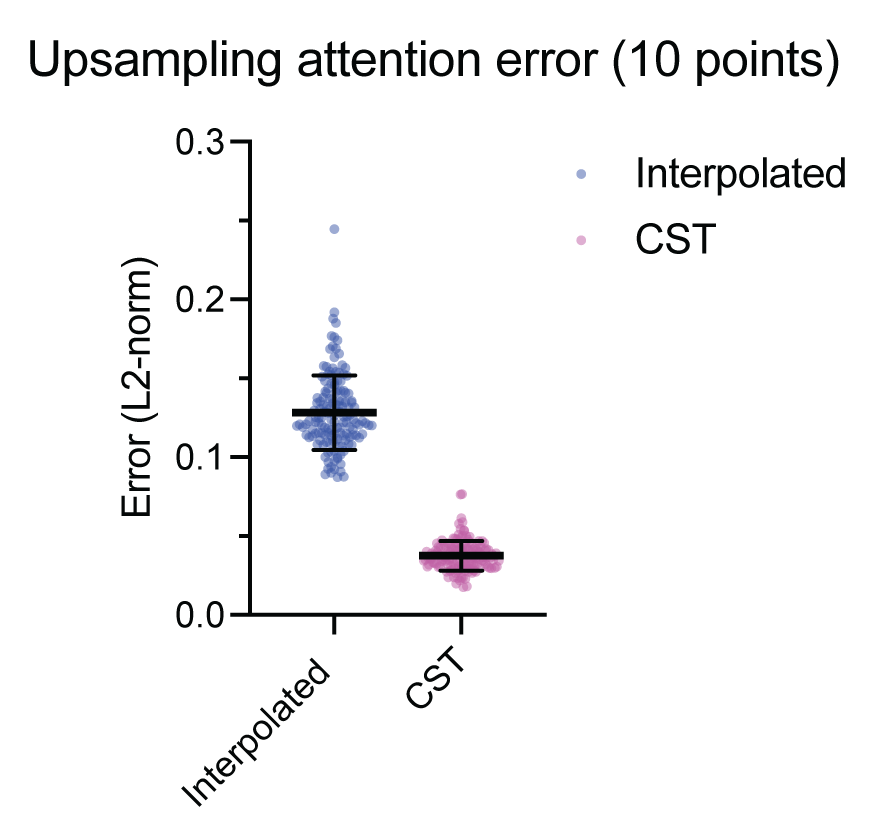}
  \caption{Distribution of error (L2-norm) for the up-sampled attention via CST vs. interpolation. (see Sec. \ref{sec:2d_spirals}).}
  \label{fig:CST_SelfAttention_Upsampling_Error}
\end{figure}



\begin{table}
\caption{Interpolation error on the 2D IE spirals (mean $\pm$ std)}
\label{tab:interpolation_error}
\vskip 0.15in
\begin{center}
\begin{small}
\begin{sc}
\begin{tabular}{cc}
\toprule
 Transformer & CST \\
\midrule
0.04329 $\pm$ 0.007511  & 0.03859 $\pm$ 0.01013 \\
\bottomrule
\end{tabular}
\end{sc}
\end{small}
\end{center}
\vskip -0.1in
\end{table}



\begin{figure*}[t]
\centering
  \includegraphics[width=0.9\textwidth]{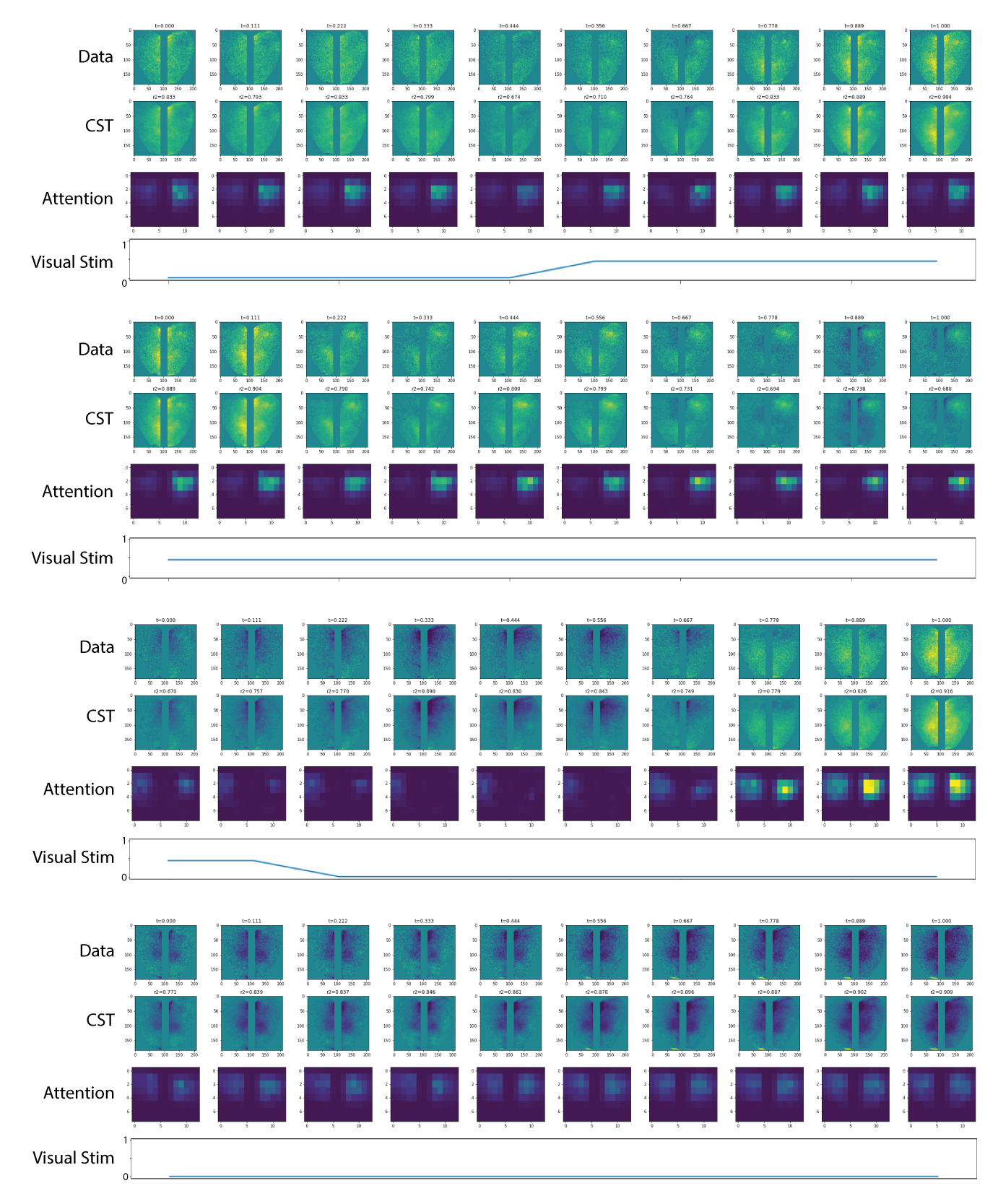}
  \caption{Examples of segments from the calcium imaging recording (see Sec. \ref{sec:CaImg})}
  \label{fig:CalImg_Attention}
\end{figure*}

\begin{table}[ht]
\caption{List of parameters for the 2D spirals experiment}
\label{tab:list_of_parameters_2DCurves}
\vskip 0.15in
\begin{center}
\begin{small}
\begin{sc}
\begin{tabular}{ccc}
\toprule
& Architecture & N. of parameters \\
\midrule
Transformer & layers=4, heads=4, dim=32, ff=3072  & 818 K  \\
CST & layers=4, heads=4, dim=32, ff=3072 & 818 K \\
\bottomrule
\end{tabular}
\end{sc}
\end{small}
\end{center}
\vskip -0.1in
\end{table}

\begin{table}[ht]
\caption{List of parameters for the KITTI experiment} 
\label{tab:list_of_parameters_KITTI}
\vskip 0.15in
\begin{center}
\begin{small}
\begin{sc}
\begin{tabular}{cp{11cm}c}
\toprule
&Architecture & N of parameters \\
\midrule
ConvGRU &  Enc: (2 ConvGRUCell, hid=[64,128]), \newline Dec: (2 ConvGRUCell, hid=[128,256]) &  53.6M\\
3DViT & Enc: Transformer: (4 layers,  heads=6, dim=1024, Spatial patch=[16,32], Temporal patch=2), \newline Dec: (1 FC layer, dim=3072) & 214.6 M\\
ViViT & Enc: (1 FC layer, dim=1024), \newline Spatial Transformer: (2 layers,  heads=6, dim=768, patches=[16,32]), \newline Temporal Transformer (2 layers,  heads=6, dim=768, patches=2), \newline Dec: (1 FC layer, dim=3072) & 224.7 M\\
CST & Enc: (1 FC layer, dim=768), \newline Transformer (layers=12, heads=12, dim=768, ff=3072), \newline Dec: (1 FC layer, dim=768) & 93.3M \\
\bottomrule
\end{tabular}
\end{sc}
\end{small}
\end{center}
\vskip -0.1in
\end{table}

\begin{table}[ht]
\caption{List of parameters for the Navier–Stokes experiment}
\label{tab:list_of_parameters_NS}
\vskip 0.15in
\begin{center}
\begin{small}
\begin{sc}
\begin{tabular}{cp{11cm}c}
\toprule
& Architecture & N of parameters \\
\midrule
FNO2D & Enc: (4 SpectralConv2D layers, dim=20) \newline 
Forw: (4 CNN2D, dim=[20,20,20,20]) \newline
Dec: (2 FC layers, dim=[20,128]) & 926 K  \\
FNO3D & Enc: (4 SpectralConv3D layers, dim=20) \newline 
Forw: (8 CNN3D, dim=[20,...,20]) \newline
Dec: (2 CNN3D, dim=[20,80]) & 6.56 M   \\
Transformer & layers=4, heads=4, dim=32, FF=3072 & 818 K \\
CST & layers=4, heads=4, dim=32, FF=3072 & 818 K \\
\bottomrule
\end{tabular}
\end{sc}
\end{small}
\end{center}
\vskip -0.1in
\end{table}

\begin{table}[ht]
\caption{List of parameters for the calcium imaging experiments}
\label{tab:list_of_parameters_CaImg}
\vskip 0.15in
\begin{center}
\begin{small}
\begin{sc}
\begin{tabular}{cp{11cm}c}
\toprule
& Architecture & N. of parameters \\
\midrule
LatentODE & ODE func: (3 layers,dim=40), \newline Rec. RNN: (2 layers,hid=25), \newline Dec: (2 layers,dim=40)  & 2.5M  \\
LSTM & Enc: (1 LSTMcell, hid=240), \newline Dec: (1 FC layer,dim=240)  & 46.2M \\
ConvLSTM &  Enc: (2 ConvLSTMCell, hid=256), \newline Dec: (2 ConvLSTMCell, hid=256, 3DCNN) &  16.5M\\
ViViT & Enc: (1 FC layer, dim=1024), \newline Spatial Transformer: (2 layers,  heads=6, dim=768, patches=64), \newline Temporal Transformer (2 layers,  heads=6, dim=768, patches=64), \newline Dec: (1 FC layer, dim=3072) & 528.9 M\\
LFADS & Enc generator: (forw.: 1 GRUcell, hid=200, back.: 1 GRUcell, hid=200) \newline 
Enc controller: (forw.: 1 GRUcell, hid=128, back.: 1 GRUcell, hid=128) \newline
controller: (1 GRUcell, hid=128), generator: (1 GRUcell, hid=200) \newline
Factors: (1 FC layer, dim=100) & 80.0M \\
CST-small & Enc: (1 FC layer, dim=384), \newline Transformer (layers=6, heads=6, dim=384, ff=3072), \newline Dec: (1 FC layer, dim=384) & 19.1M \\
CST-base & Enc: (1 FC layer, dim=768), \newline Transformer (layers=12, heads=12, dim=768, ff=3072), \newline Dec: (1 FC layer, dim=768) & 93.3M \\
\bottomrule
\end{tabular}
\end{sc}
\end{small}
\end{center}
\vskip -0.1in
\end{table}

\begin{figure*}[t]
\centering
  \includegraphics[width=0.9\textwidth]{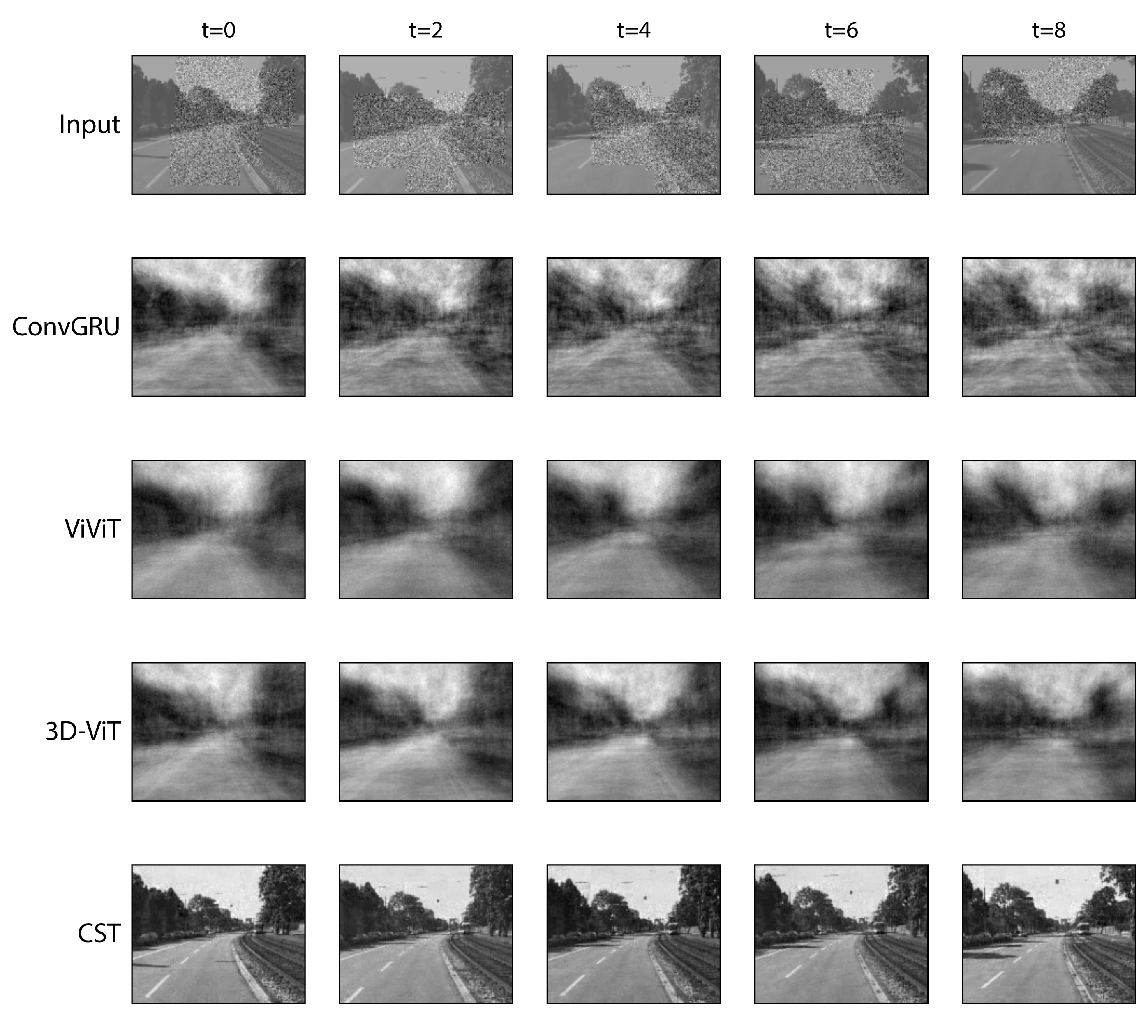}
  \caption{Top: Validation set example sequence from KITTI Dataset, and prediction from all models trained on the Video inpainting task.}
  \label{fig:example_visualization_video_cst}
\end{figure*}

\begin{figure*}[t]
\centering
  \includegraphics[width=0.5\textwidth]{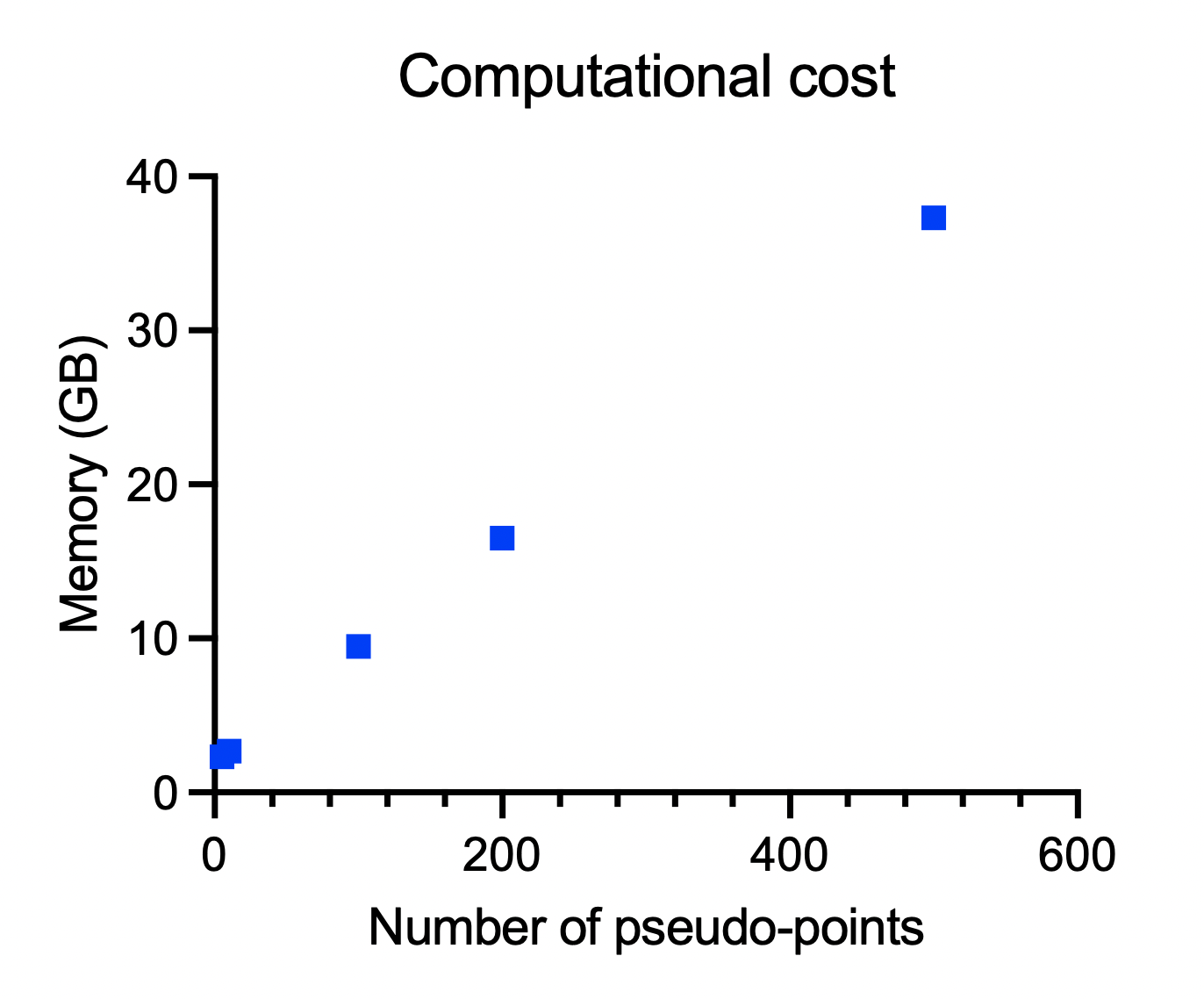}
  \caption{Computational cost expressed in terms of GPU memory usage for a given number of dummy-points added during training.}
  \label{fig:computational_cost}
\end{figure*}

      
\end{document}